\pdfoutput=1

\documentclass[11pt]{article}

\usepackage{wrapfig}

\usepackage{mathpazo}
\usepackage[utf8]{inputenc}
\usepackage{verbatim}
\usepackage{calc}
\usepackage{mathrsfs}
\usepackage{mathtools}
\usepackage[hyphens]{url}

\usepackage{amsmath}
\usepackage{amssymb}
\usepackage{graphicx}
\usepackage{hyperref}
\usepackage{xcolor}
\usepackage{comment}
\usepackage{soul}
\usepackage{booktabs}

\definecolor{forestgreen}{rgb}{0.13, 0.55, 0.13}

\usepackage{tabularx}
\usepackage{multirow}
\usepackage{multicol}

\usepackage{colortbl}
\usepackage{tabulary}
\usepackage{etoolbox}
\usepackage{enumitem}

\newcommand\hb{ \rowcolor{teal!4}}
\newcommand\hc{ \rowcolor{teal!10}}
\newcommand\hd{ \rowcolor{teal!18}}

\definecolor{colorA}{RGB}{189,201,225}
\definecolor{colorB}{RGB}{103,169,207}
\definecolor{colorC}{RGB}{ 28,144,153}
\definecolor{colorD}{RGB}{  1,108, 89}

\newcolumntype{R}{>{\columncolor{gray!40}}r}
\newcolumntype{L}{>{\columncolor{gray!40}}l}
\newcolumntype{C}{>{\columncolor{gray!40}}c}

\usepackage[many]{tcolorbox}
\usepackage{caption}
\usepackage{subcaption}
\usepackage{graphbox} 

\tcbset{
    sharp corners,
    colback = white,
    before skip = 0.2cm,    
    after skip = 0.5cm      
}                           

\definecolor{main}{HTML}{4472C4}    
\definecolor{sub}{HTML}{EBF4FF}     

\newtcolorbox{boxA}{
    enhanced, breakable,
    boxrule = 0pt,
    colback = sub,
    borderline west = {2pt}{0pt}{main}, 
    borderline east = {2pt}{0pt}{main}, 
}

\usepackage[defaultlines=2,all]{nowidow}

\usepackage[final]{acl}

\usepackage{times}
\usepackage{latexsym}

\usepackage[T1]{fontenc}

\usepackage[utf8]{inputenc}

\usepackage{microtype}

\usepackage{inconsolata}

\usepackage{graphicx}
\usepackage{xspace}

\newcommand{\OURS}{Squeezed Attention\xspace}
\newcommand{\BENCH}{{PreFixQA}\xspace}

\title{\OURS: Accelerating Long Context Length LLM Inference}

\author{
 \textbf{Coleman Hooper\thanks{Equal contribution}\textsuperscript{1}},
 \textbf{Sehoon Kim\footnotemark[1]\textsuperscript{1}},
 \textbf{Hiva Mohammadzadeh\textsuperscript{1}},
 \\
 \textbf{Monishwaran Maheswaran\textsuperscript{1}},
 \textbf{Sebastian Zhao\textsuperscript{1}},
 \textbf{June Paik\textsuperscript{2}},
 \\
 \textbf{Michael W. Mahoney\textsuperscript{1,3,4}},
 \textbf{Kurt Keutzer\textsuperscript{1}},
 \textbf{Amir Gholami\textsuperscript{1,3}}
 \\
 \textsuperscript{1}UC Berkeley,
 \textsuperscript{2}FuriosaAI,
 \textsuperscript{3}ICSI,
 \textsuperscript{4}LBNL
 \\
 \small{
 \textbf{Correspondence to:} chooper@berkeley.edu
 }
}

\begin{document}
\maketitle
\begin{abstract}

Emerging Large Language Model (LLM) applications require long input context in order to perform complex tasks like document analysis and code generation.
For these long context length applications, the length of the input prompt poses a significant challenge in terms of inference efficiency since the inference costs increase linearly with sequence length.
However, for many of these applications, much of the context in the prompt is fixed across different user inputs, thereby providing the opportunity to perform offline optimizations in order to process user inputs quickly, as they are received. 
We propose \textit{\OURS} to accelerate LLM applications where a large portion of the input context is fixed.
We first leverage K-means clustering offline to group the keys for the fixed context based on semantic similarity and represent each cluster with a single centroid value.
During inference, we compare query tokens from the user input with the centroids to predict which keys from the fixed context are semantically relevant, and then compute exact attention using only the important keys, thereby reducing bandwidth and computational costs. 
We also present a hierarchical version of our algorithm which can reduce the complexity of attention from linear to logarithmic with respect to the fixed context length.
We evaluate our method on various long-context benchmarks including LongBench, where it achieves a 3.1$\times$ reduction in KV budget with no noticeable accuracy loss and up to an 8$\times$ reduction with only a 0.5 point accuracy gap for the LLaMA-2-7B-32K, LWM-Text-Chat-1M, and Longchat-7B-v1.5-32K models.
Futhermore, we implement kernels for centroid comparison and sparse FlashAttention with important keys, achieving more than 4$\times$ speedups during both the prefill and generation phases for long-context inference.
Our code is available at 
\url{https://github.com/SqueezeAILab/SqueezedAttention}.
\end{abstract}

\section{Introduction}

\begin{figure}[t]
\centering
\includegraphics[width=\linewidth]{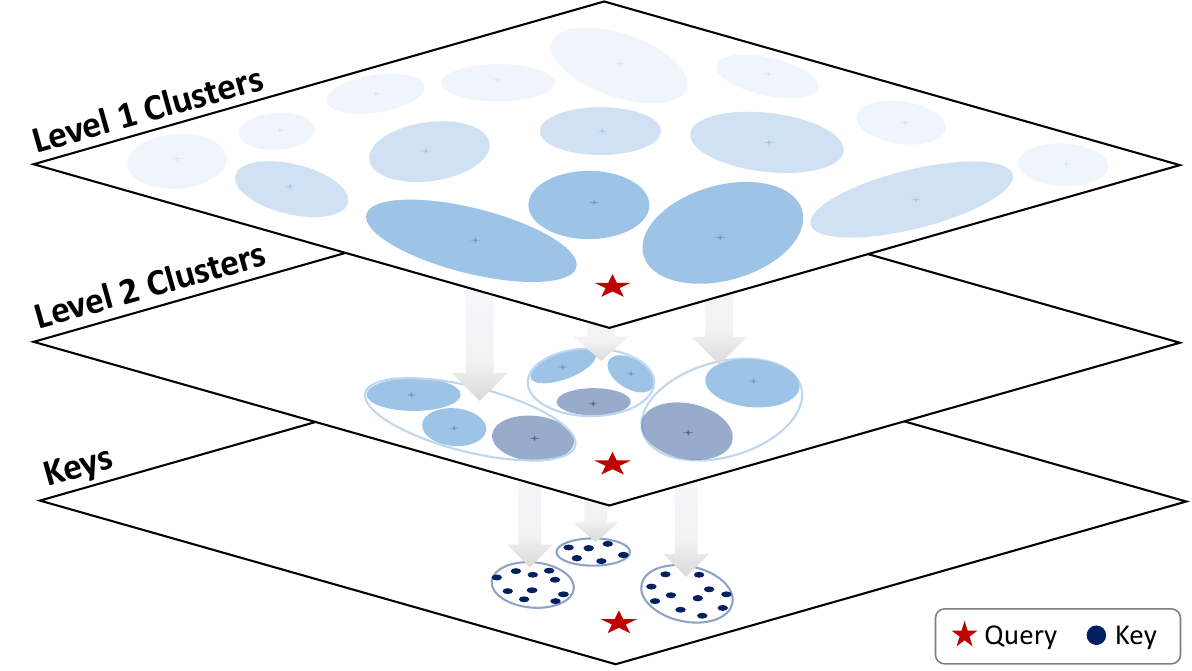}
\vspace{-5mm}
 \caption{
 A high-level visualization of our hierarchical clustering approach. 
 We identify important keys for the current query by first identifying which coarse-grained clusters are relevant (Level 1). 
 We then refine this prediction using finer-grained clustering (Level 2). 
 Finally, we identify the important keys for the current query and only compute exact attention with these keys.
}
  \label{fig:thumbnail}
  \vspace{-6mm}
\end{figure}

Large Language Models (LLMs) have seen rapid advancements in recent years, enabling a range of downstream applications including Question Answering (QA) and analysis over structured and unstructured documents.
Performance on these tasks has benefited from the increased context lengths of newer open-source \cite{liu2024world, touvron2023llama} and closed-source \cite{achiam2023gpt, claude2, gemini15} models, as these tasks benefit from incorporating a large amount of input context in order to condition the model to generate particular outputs. 
However, deployment of LLMs for downstream applications is constrained by inference costs, with LLM inference requiring significant computational resources as well as memory capacity and bandwidth.
In particular, long context-length applications have large memory capacity and memory bandwidth requirements due to the size of the KV cache, which increases linearly with respect to sequence length \cite{tang2024quest,hooper2024kvquant,li2024snapkv}.

For many applications such as in-context learning, document QA, and code generation, over a series of prompts a large portion of the input context is fixed.
This fixed context, which may contain system instructions,
documentation, or few-shot examples, is extremely beneficial for tailoring the model to the target application.
However, increasing the length of the fixed context poses a significant challenge for inference efficiency. 
Throughout this work, we will refer to this portion of the prompt as the ``fixed context,'' and we will refer to the portion that corresponds to the user requests that come in online as the ``user input.''
The user input is appended after the fixed context and provided to the model. 
For many long-context applications, the fixed context portion of the prompt is much longer than the user input portion of the prompt, and the attention computation for this fixed context portion typically dominates the inference runtime.
In this work, our aim is to take advantage of the fact that this context is fixed and available prior to inference in order to optimize attention to the fixed context for new user requests.

We propose \textit{\OURS} as a method to accelerate fixed context applications by accelerating the attention computation.
Our method, illustrated in Figure \ref{fig:thumbnail}, accelerates inference by quickly identifying which keys in the fixed context are important for a given query token.
Offline prior to inference, we cluster the keys in the fixed context based on their semantic similarity and then represent keys from the same cluster using a single representative ``key centroid''.
At inference time, when the user input is received, we retrieve the important keys by first comparing the query tokens with the key centroids, rather than the entire set of keys, in order to identify the important key clusters for the current query. 
Once the important clusters are identified, we retrieve their associated keys and compute exact attention only with those high-scoring keys.
Our method can be further extended to a \textit{hierarchical} clustering and retrieval scheme, as shown in Figure~\ref{fig:thumbnail}, efficiently narrowing the search space by first leveraging coarser-grained clusters and then refining the search using fine-grained clusters.
In contrast to existing solutions~\cite{zhang2024h2o,li2024snapkv,ge2023model}  that identify less important tokens once and discard them throughout the entire generation, our method \textit{dynamically} identifies and retrieves only the information that is \textit{semantically} relevant to each generation step.
This allows our method to preserve generation quality while reducing the number of KV cache entries loaded from memory by up to 8 times (including loading key centroids), as highlighted in Section~\ref{sec:results}.
By optimizing memory bandwidth as well as computational costs, \OURS effectively reduces overheads for both generation and prefill during long-context inference. 

Our work makes the following contributions:
\vspace{-2.5mm}
\begin{itemize}[leftmargin=3mm]
    \item \textbf{Semantic-based Key Clustering and Retrieval:} 
    To cluster non-consecutive keys by their semantic similarity, we perform K-means clustering offline, representing all keys within each cluster with a single ``key centroid'' value (Section~\ref{sec:method-cluster}). 
    This allows us to identify semantically relevant keys for the query tokens during inference by comparing the query against key clusters instead of the entire key set (Section~\ref{sec:method-online}), and only performing exact attention computation with the most relevant keys.
    Since the number of key centroids is significantly smaller than the number of keys, the memory overhead remains minimal. 
    We additionally propose a hierarchical version of our method (outlined in Section \ref{subsec:hierarchical_lookup}), which can reduce the memory and computational complexity of the centroid lookup from linear to logarithmic with respect to the fixed context length.
    \vspace{-2.5mm}
    \item \textbf{System Implementation:}
    We design efficient Triton kernels for performing the centroid comparison (Section~\ref{subsec:centroid-lookup}) and computing sparse FlashAttention with only the important keys (Section~\ref{subsec:sparse-flash-attn}).
    Combined together, our method results in 4.3$\times$ and 4.2$\times$ speedups during the prefill and decode phases when running inference with long fixed context (Section~\ref{sec:system_results}).
    \vspace{-2.5mm}
    \item \textbf{Benchmark:} 
    We introduce \BENCH (Section \ref{sec:bench}), a document QA benchmark which contains a selection of arXiv documents, each with many synthetic user input question and answer pairs.
    This benchmark facilitates research into fixed context methods by allowing us to evaluate various user inputs for each document.
    \vspace{-2.5mm}
    \item \textbf{Evaluation:} 
    We extensively evaluate our method on long-context benchmarks including LongBench~\cite{bai2023longbench}, RULER~\cite{hsieh2024ruler}, and \BENCH (Section \ref{subsec:eval}).
    On LongBench, our method preserves accuracy with 3.1$\times$ KV budget reduction and achieves up to $8\times$ KV budget reduction with 0.5 point accuracy degradation for the 
    LLaMA-2-7B-32K, LWM-Text-Chat-1M, and Longchat-7B-v1.5-32K models.
\end{itemize}

\begin{figure*}[t]
\centering
\includegraphics[width=1\linewidth]{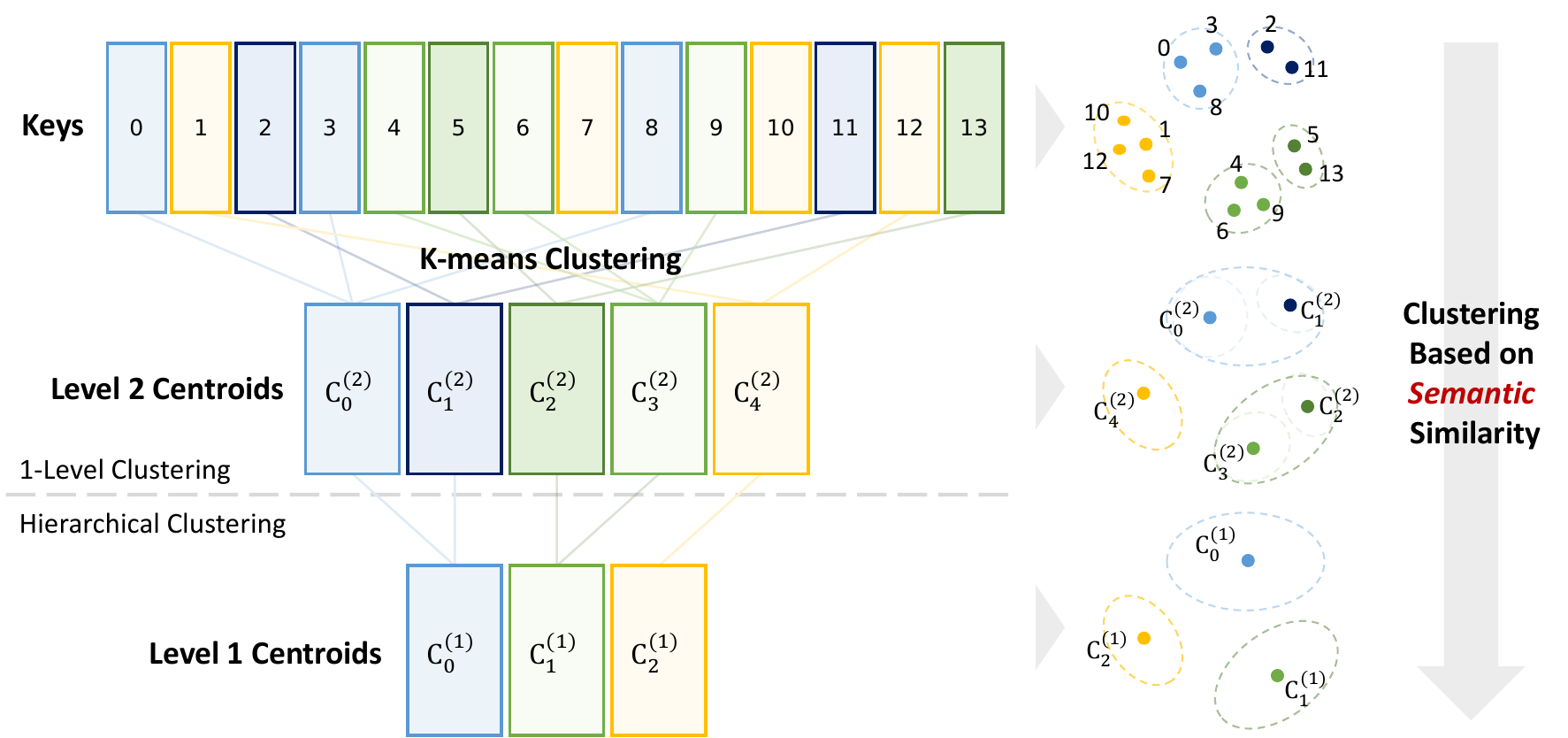}
\vspace*{-4mm}
\caption{
Diagram outlining our approach for performing clustering offline with the fixed context. Refer to Section~\ref{sec:method-cluster} for 1-level clustering and Section~\ref{subsec:hierarchical_lookup} for hierarchical clustering.
We apply K-means clustering to group semantically similar key tokens,
assigning a single centroid to represent each cluster. 
In the hierarchical approach (Section~\ref{subsec:hierarchical_lookup}, demonstrating a 2-level hierarchy for clarity),
these centroids form the Level 2 centroids, which are then clustered into coarser-grained Level 1 centroids by repeating the same procedure.
}
\label{fig:offline-clustering}
\end{figure*}

\section{Related Work}

With the growing popularity of long-context applications, there has been a continuous development of LLMs that can support context lengths exceeding 100k, and even up to 1M tokens \cite{achiam2023gpt,claude2,gemini15, liu2024world}.
However, as context lengths increase, the KV cache often becomes a critical bottleneck, significantly impacting memory usage and latency during LLM inference~\cite{tang2024quest,hooper2024kvquant}.
Therefore, KV cache compression methods have emerged as a critical approach for enabling efficient inference when using long-context models.
We provide a brief summary here of relevant KV cache compression works; an extended discussion of related work for long-context length inference and KV compression is provided in Appendix \ref{sec:appendix-related-work}.

Several previous methods have been proposed to enable more efficient long-context inference by reducing the KV cache size \cite{hooper2024kvquant,liu2024kivi,fu2024lazyllm,zhang2024h2o}.
A notable approach is KV cache sparsification, which encompasses two general directions: KV cache eviction, and sparsely loading the KV cache. 
KV eviction compresses the KV cache by identifying and removing less important tokens \cite{zhang2024h2o,oren2024transformers,li2024snapkv}.
While these methods are effective for reducing memory requirements, they cannot recall evicted tokens later during generation if these become important, and they therefore cannot adapt to changing token importance during generation or in response to subsequent user inputs.
Another direction which is most similar to our approach is sparse KV cache loading, where the full KV cache is stored but only relevant keys and values are loaded dynamically during inference. 
One prior work, QUEST~\cite{tang2024quest}, clusters consecutive KV cache entries and dynamically retrieves the most relevant clusters based on their relevance to each query token during generation.
However, this approach relies on \textit{physical} proximity for clustering, whereas clustering should instead be based on \textit{semantic} proximity, as tokens that are physically far apart can be semantically similar. 
\OURS addresses this by clustering tokens based on embedding similarity, ensuring that semantically relevant tokens are retrieved for future generations.

\begin{figure*}[t]
\centering
\includegraphics[width=\linewidth]{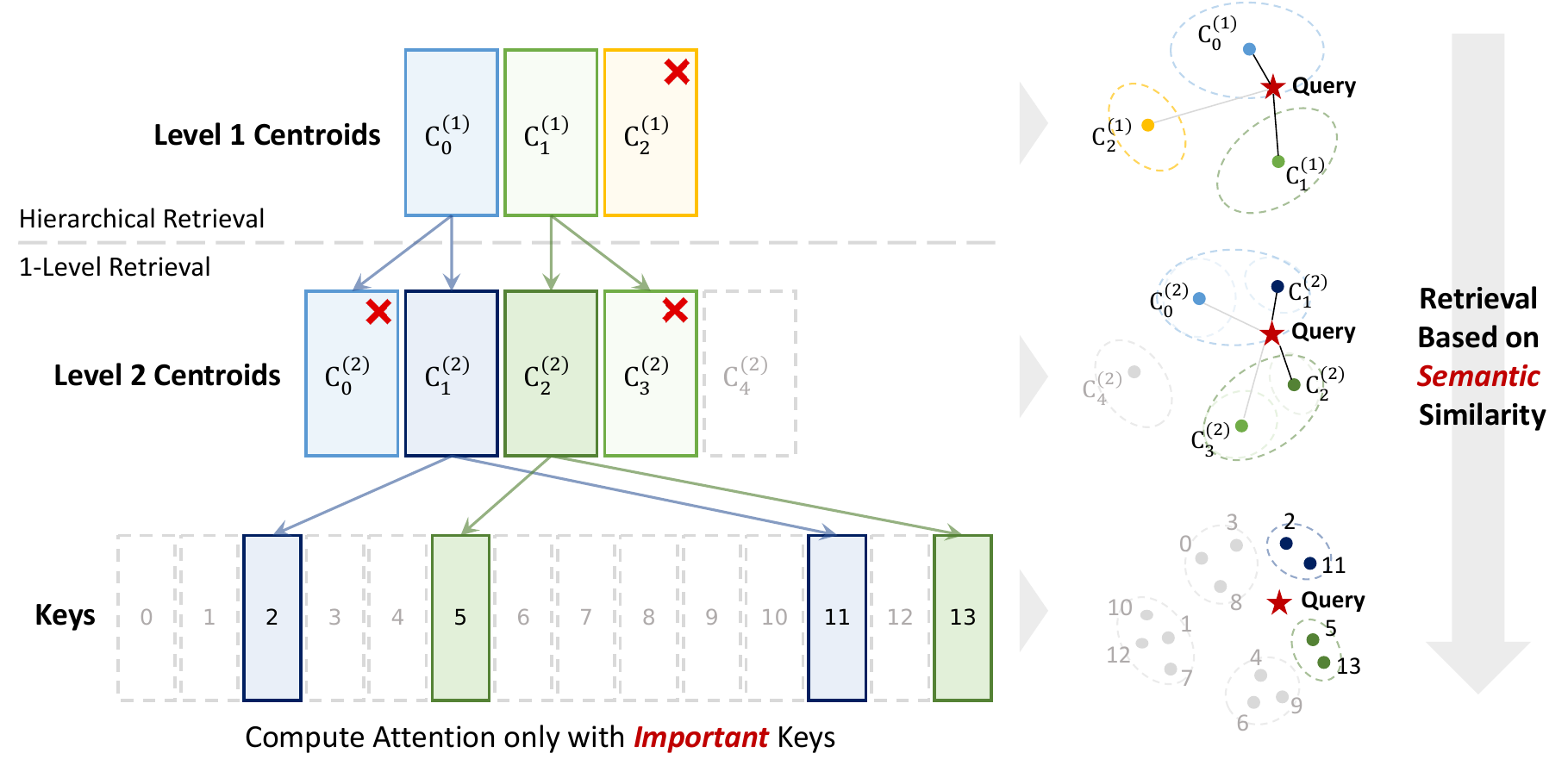}
\vspace*{-4mm}
 \caption{
Diagram outlining how our method operates during inference to retrieve the most relevant keys when a new input query is received.
Refer to Section~\ref{sec:method-online} for 1-level retrieval and Section~\ref{subsec:hierarchical_lookup} for hierarchical retrieval.
For 1-level retrieval, the query token is first compared against the representative centroid of each cluster to identify the most relevant clusters. 
Exact attention is then computed only for the keys within these retrieved clusters, rather than across the entire fixed context.
In our hierarchical retrieval approach (Section~\ref{subsec:hierarchical_lookup}, demonstrating a 2-level hierarchy for clarity), we first compare the query with coarse-grained Level 1 centroids, and then only compare with a subset of the promising fine-grained Level 2 centroids in order to identify the important keys.
}
  \label{fig:online-lookup}
\end{figure*}

\section{Algorithm}
\label{sec:algorithm}

\subsection{Offline: Clustering Keys}
\label{sec:method-cluster}

The first step in our method is to cluster the fixed context keys offline based on \textit{semantic} similarity, as outlined in Figure~\ref{fig:offline-clustering}.
Specifically, we use K-means clustering with normalized key vectors to group similar keys together.
We then take the mean of all vectors in each cluster to obtain a representative centroid, which can be used as a representative key token for all tokens in that cluster. 
By comparing incoming queries with this centroid, we can determine whether the tokens in that cluster are important without necessarily comparing them with individual keys.
Note that this \textit{semantic-based} clustering approach groups together non-consecutive key tokens, which could make it more challenging to efficiently load the keys from memory.
However, the size of each KV cache token for a single head in modern LLMs is typically at least 256 bytes in bf16 (as the head dimensions are typically at least 128) \cite{touvron2023llama,touvron2023llama2}, which is sufficiently large to efficiently utilize memory bandwidth.
Therefore, we are still able to execute memory operations efficiently when sparsely loading in the non-consecutive keys and associated values from memory.

\subsection{Online: Query-Aware Key Retrieval}
\label{sec:method-online}

During inference, we would ideally only load the keys which would have high attention scores for the current query, but this cannot be known ahead of time without doing a full pass over the keys.
We leverage the cluster centroid to approximately measure the ``average'' attention score of the keys within a cluster, thereby allowing us to identify important keys without loading them all.
By organizing keys into clusters, each represented by a single centroid, we can accelerate inference for incoming user inputs, as shown in Figure~\ref{fig:online-lookup}.
We first compare the input query tokens with the key centroids to assess which key tokens are likely to have high attention scores. 
We estimate the importance of cluster $i$ for query token $q$ as:

\vspace{-7.5pt}
\begin{equation}
\label{eq:centroid_lookup}
    S_i = \frac{\exp{(q  C_{i}^\top)}}{\sum_j N_j \cdot \exp{(q  C_{j}^\top)} },
\end{equation}
\vspace{-7.5pt}

\noindent
where $N_j$ is the number of keys in cluster $j$ and $C_j$ is the cluster centroid for cluster $j$. 
This allows us to assess the average importance of the tokens in cluster $i$. 
If the average importance of a cluster is above a desired threshold, we load in the keys for that cluster and perform exact attention computation; otherwise, we avoid loading and performing computation with these keys.

Using the Softmax estimate $S_i$, instead of $qC_i^\top$, as an importance metric for each cluster provides an easy method to control the number of important keys retrieved from each attention head.
As outlined in Appendix \ref{appendix:skewness}, some attention heads have a more balanced distribution of attention scores, resulting in a larger number of important keys, while others have a more skewed distribution, indicating only a few important keys. 
Ideally, we want to retrieve more keys from heads with a larger number of important keys.
Since Softmax values are normalized to sum to 1, we can apply a single \textit{global} threshold across all layers and attention heads to achieve this. 
This allows us to automatically retrieve more keys from heads with balanced attention score distributions, where more $S_i$ values exceed the threshold; and fewer keys from heads with skewed distributions, where fewer $S_i$ values exceed the threshold.
This approach eliminates the need for manually configuring the number of keys to retrieve for each head.
Once we choose the threshold to achieve the desired sparsity level, it is kept throughout the prefill and generation stages.
Details about the calibration procedure for the global threshold are provided in Appendix \ref{appendix:calibration}.

\subsection{Hierarchical Centroid Lookup}
\label{subsec:hierarchical_lookup}
The centroid lookup approach outlined in Sections \ref{sec:method-cluster} and \ref{sec:method-online} quickly identifies important keys for the attention computation and only computes attention for these keys.
As long as we use fine-grained centroids, we can have sufficient resolution to identify which keys will be important and we can retain accuracy.
However, it is desirable to have fewer centroids since a larger number of centroids leads to an increased cost for centroid lookup.

In order to attain the accuracy improvements of fine-grained centroid lookup while retaining the efficiency benefits of using coarse-grained centroids, we leverage a \textit{hierarchical} centroid lookup process.
Figure \ref{fig:offline-clustering} demonstrates the offline preprocessing step with our (two-level) hierarchical approach.
Initially, using the same approach as in Section~\ref{sec:method-cluster}, we cluster the keys into a larger number of centroids, referred to as Level 2 centroids. 
We then perform K-means clustering on these Level 2 centroids to produce a smaller number of coarse-grained centroids, referred to as Level 1 centroids.

During inference, we perform a hierarchical centroid lookup as outlined in Figure \ref{fig:online-lookup}.
We first compare incoming queries with the coarse-grained Level 1 centroids to quickly prune out unnecessary keys.
This initial lookup narrows down the search, allowing us to focus on comparing the queries with the fine-grained Level 2 centroids that are likely to be high-scoring.
Specifically, we first compare the input query token $q$ with each of the coarse-grained key centroids $C^{(1)}_i$ to assess which key tokens are likely to have high attention scores:

\vspace{-10pt}
\begin{equation}
\label{eq:centroid_lookup_hier_1}
    S^{(1)}_i = \frac{\exp{(q C_{i}^{(1)\top})}}{\sum_j N^{(1)}_j \cdot \exp{(q C_{j}^{(1)\top})} }.
\end{equation}
\vspace{-10pt}

We then apply a threshold $T_1$ to rule out low-scoring clusters at the coarse-grained level, thereby avoiding comparisons with the fine-grained Level 2 centroids for the less relevant clusters. 
For the remaining Level 1 centroids, we expand them into their corresponding finer-grained Level 2 centroids, $C_l^{(2)}$, which are then compared with the input query token $q$ to assess their relevance:

\vspace{-10pt}
\begin{equation}
\label{eq:centroid_lookup_hier_2}
    S^{(2)}_l = \frac{\exp{(q C_{l}^{(2)\top})}}{\sum_m N_m ^{(2)}\cdot \exp{(q C_{m}^{(2)\top})} }.
\end{equation}
\vspace{-10pt}

Since we are only considering the remaining Level 2 centroids, the denominator is also calculated based on these selected centroids.
We then compare $S^{(2)}_l$ with threshold $T_2$ to decide which keys should be used for exact attention computation.
With this hierarchical approach, we can reduce the cost of finer-grained centroid lookup while maintaining its accuracy.
Although we describe a 2-level process here for clarity, this method can be extended to multiple levels of~hierarchy.

\subsection{Complexity Analysis}
\label{subsec:complexity}

\begin{table}[t]
\centering
\caption{Theoretical memory and compute complexity of the baseline (standard autoregressive generation), 1-level retrieval, and hierarchical retrieval for a single generation iteration. Here, $L$ represents the context length, $c$ is the number of clusters in the 1-level retrieval approach, and $k \ll L$ is the number of keys remaining after retrieval. In the hierarchical retrieval approach, $c' \ll c < L$ denotes the number of clusters at each hierarchical level.
Note that $c$ for the 1-level retrieval cannot be reduced significantly with respect to $L$, while $c'$ for the hierarchical retrieval can.}
\label{tab:complexity-analysis}
\vspace{-2mm}
\small
\begin{tabular}{cc}
\toprule  
\rotatebox{0}{\textbf{{Method}}} & 
\rotatebox{0}{\textbf{{Memory / Compute Complexity}}} \\
\midrule
Baseline & $O(L)$ \\
\midrule
1-Level  & $O(c + k)$ where $k \ll L$ \\
Hierarchical  & $O(c'\log L + k)$ where $c', k \ll L$\\
\bottomrule
\end{tabular}
\end{table}

Let $L$ denote the context length, which can be substantially large in long-prompt applications. In the baseline approach (i.e., standard autoregressive generation), each generative step requires comparing a query token with the entire set of keys in the prompt, resulting in $O(L)$ memory and compute operations per iteration (i.e. per token generation).
If we apply 1-level retrieval, however, we can instead use $c$ centroids to identify the relevant key clusters and then compute attention using only $k \ll L$ retrieved keys. 
This reduces the memory and compute complexity to $O(c + k)$ per iteration.
One limitation of the 1-level retrieval approach is that it can be challenging to significantly reduce $c$ (the number of centroids), 
as it would require clustering a large number of keys into each cluster. 
This may result in either pruning keys too aggressively or retrieving irrelevant keys grouped together in the same cluster.

In contrast, hierarchical centroid retrieval allows for a more efficient reduction in centroids at each level of the hierarchy by enabling gradual pruning of keys. 
Suppose we retrieve only $c' \ll c < L$ clusters at each hierarchical level.
In this setup, we require $O(\log L)$ hierarchical levels to reduce the keys to the desired final count, $k$.
Therefore, as highlighted in Table \ref{tab:complexity-analysis}, the memory and compute complexity for each generation iteration becomes $O(c' \log L + k)$, reducing the complexity from \textit{linear to logarithmic} with respect to the context length.

\section{System Implementation}

\subsection{Centroid Lookup}
\label{subsec:centroid-lookup}

The first stage of our kernel implementation compares query tokens with the centroids for the fixed context keys. 
These query tokens may include multiple tokens from an incoming user input during the prefill stage or a single token during the generation stage.
The kernel follows similar parallelization strategies to FlashAttention-2 \cite{dao2023flashattention}, where we split across different attention heads and along the query sequence length dimension. 
We first load a block of query tokens and iterate over the entire key centroids in order to find the most important key centroids according to Equation~\ref{eq:centroid_lookup}.
At a high level, the kernel performs an initial pass over the key centroids to compute the denominator in Equation~\ref{eq:centroid_lookup} based on the query-key centroid dot product.
Then, it takes a second pass over the centroids to compute $S_i$ as in Equation~\ref{eq:centroid_lookup}, using the denominator results from the first pass. 
Finally, we compare $S_i$ with a target threshold $T$, and we only load the keys in cluster $i$ if $S_i > T$. 

During prefill, where multiple query tokens are available, we split the workload along the query sequence length dimension as in FlashAttention-2 \cite{dao2023flashattention} to attain additional parallelism.
During generation, achieving parallelism is more challenging, as we cannot leverage parallelism across the dimension of the length of the query sequence. 
This is particularly problematic when dealing with small batch sizes, as in that case the only parallelism we can leverage is across different heads.
To address this, we develop an optimized implementation which parallelizes the second pass over the centroids across the cluster dimension. Appendix \ref{appendix:kernel-ablation} highlights how this optimized implementation is critical for performing the centroid lookup during generation without substantial latency overhead.
Additional details of how the centroid lookup is applied during the prefill and generation stages (as well as for the kernel implementation for our hierarchical method) are provided in Appendix \ref{appendix:centroid-lookup}.

\subsection{Sparse Attention with Retrieved Keys}
\label{subsec:sparse-flash-attn}

Once the important keys are identified through our centroid lookup, the second stage of our system implementation leverages a sparse FlashAttention kernel to attain speedups during \textit{both} prefill and generation stages.
This stage also uses similar parallelization strategies as FlashAttention-2 by splitting work across heads and along the sequence length dimension \cite{dao2023flashattention}.
Our kernel implementation builds on top of prior work on Triton implementations for FlashAttention-2 
\cite{flashattentiontutorial} and for dynamic sparse FlashAttention \cite{pagliardini2023fast}.
The kernel uses a tensor containing key indices to identify which keys need to be loaded from memory for exact attention computation.

An additional challenge when computing attention to the fixed context is the imbalanced distribution of important key tokens across different heads, which is highlighted in Appendix~\ref{appendix:skewness}.
When using the default parallelization strategy in FlashAttention-2, if one head contains more important keys than the other heads, it will have significantly longer runtime, hindering speedups. 
In order to obtain latency benefits in these scenarios, we split keys and values along the sequence-length dimension as in Flash-Decoding \cite{flashdecoding}, based on a fixed number of desired keys and values to be computed for a single Streaming Multiprocessor (SM). 
This means that if there are more keys and values that need to be loaded for a particular head, the work for this head will be parallelized across a greater number of SMs in the GPU. 
As in Flash-Decoding \cite{flashdecoding}, our kernel is designed in two phases, with the first phase computing partial attention outputs and the second stage merging the partial attention outputs.

\begin{table*}[h!]
\centering
\caption{
LongBench evaluation results with \OURS. 
We report results for the Llama-2-7B-32K, LWM-Text-Chat-1M, and Longchat-7b-v1.5-32K models, using our single-level (``Sq'') and hierarchical (``H-Sq'') lookup approaches.
We also report baseline comparisons with QUEST \cite{tang2024quest}, demonstrating how our use of semantic similarity when clustering keys outperforms grouping keys sequentially.
We report the average score across LongBench tasks, as well as the average without SamSum (``All*'') for comparison against QUEST, whose evaluation framework does not support SamSum.
We also include the KV budget (``Budget''), which gives the expected percentage of the KV cache that needs to be loaded in during inference (including extra memory movement for cluster centroids).
Additional experimental details are provided in Appendix \ref{appendix:experimental}.
}
\label{tab:longbench}
\scriptsize
\setlength{\tabcolsep}{3.2pt}{
\begin{tabular}{c|c|ccc|ccc|ccc|ccc|cc|cc}
\toprule
&  \multirow{2}{*}{\raisebox{-4.8mm}{\textbf{Budget}}} & \multicolumn{3}{c|}{\textbf{Single-Doc. QA}} & \multicolumn{3}{c|}{\textbf{Multi-Doc. QA}} & \multicolumn{3}{c|}{\textbf{Summarization}} & \multicolumn{3}{c|}{\textbf{Few-shot Learning}} & \multicolumn{2}{c|}{\textbf{Code}} & \multicolumn{2}{c}{\textbf{Avg.}} \\
\cline{3-18}
\raisebox{4.2mm}{\textbf{Config}} & & \rotatebox{45}{\textbf{NQA}} & \rotatebox{45}{\textbf{Qspr}} & \rotatebox{45}{\textbf{MFQA}} & \rotatebox{45}{\textbf{HPQA}} & \rotatebox{45}{\textbf{2Wiki}} & \rotatebox{45}{\textbf{Musique}} & \rotatebox{45}{\textbf{GRep}} & \rotatebox{45}{\textbf{QMSum  }} & \rotatebox{45}{\textbf{MNews}} & \rotatebox{45}{\textbf{TREC}} & \rotatebox{45}{\textbf{TQA}} & \rotatebox{45}{\textbf{SSum}}  & \rotatebox{45}{\textbf{RB}}& \rotatebox{45}{\textbf{LCC}} & \raisebox{2.4mm}{\textbf{All*}}  & \raisebox{2.4mm}{\textbf{All}} \\ 
\midrule
\multicolumn{18}{c}{\textbf{LLaMA-2-7B-32K}} \\ 
\midrule
All KV &  1 & 17.91 & 11.12 & 33.87 & 12.45 & 11.95 & 6.54 & 29.37 & 16.93 & 21.58 & 71.50 & 87.96 & 43.87 & 61.45 & 59.14 & 33.98 & 34.69  \\
\midrule
\hc Sq-70\% & 0.325  & 18.55 & 11.78 & 34.33 & 12.31 & 12.31 & 6.26 & 29.50 & 16.90 & 20.76 & 69.00 & 87.96 & 43.90 & 61.29 & 59.53 & {33.88} & 34.60 \\
\midrule
QUEST & 0.215 & 17.01 & 9.89 & 32.10 & 11.94 & 11.41 & 6.27 & 28.90 & 17.65 & 22.14 & 68.00 & 86.43 & - & 62.53 & 59.39 & 33.36  & - \\
\hc Sq-80\% & 0.225 & 19.03 & 12.11 & 32.77 & 12.51 & 11.53 & 6.66 & 28.82 & 17.19 & 20.70 & 69.00 & 87.46 & 44.42 & 61.26 & 59.78 & \textbf{33.76} & 34.52  \\
\midrule
QUEST & 0.168 & 20.42 & 9.72 & 29.46 & 11.45 & 9.75 & 5.46 & 27.06 & 17.20 & 21.83 & 68.50 & 86.36 & - & 61.93 & 59.38 &  32.96  & - \\
\hc Sq-90\%  & 0.125 & 18.15 & 14.39 & 32.38 & 11.84 & 11.70 & 6.45 & 29.06 & 16.93 & 21.66 & 70.00 & 87.43 & 45.15 & 58.79 & 59.37 & \textbf{33.70} & 34.52 \\
\hd H-Sq-90\% & 0.112 & 17.41 & 14.23 & 32.71 & 11.99 & 11.38 & 6.68 & 29.14 & 16.97 & 20.41 & 68.00 & 87.37 & 44.85 & 58.94 & 59.61 & 33.45 & 34.26 \\
\midrule
\multicolumn{18}{c}{\textbf{LWM-Text-Chat-1M}}  \\ 
\midrule
All KV & 1  & 16.27 & 24.36 & 42.00 & 21.63 & 16.70 & 9.10 & 27.57 & 24.71 & 24.48 & 70.50 & 61.70 & 39.59 & 41.77 & 40.72 & 32.42 & 32.94\\
\midrule
\hc Sq-70\% & 0.325 & 16.54 & 24.71 & 42.24 & 21.66 & 15.88 & 9.08 & 27.28 & 24.77 & 24.60 & 70.50 & 60.93 & 39.75 & 41.06 & 40.76 & {32.31} & 32.84 \\
\midrule
QUEST & 0.215 & 15.24 & 24.57 & 40.68 & 21.57 & 17.02 & 7.93 & 27.29 & 24.86 & 24.45 & 67.00 & 62.14 & - & 45.53 & 43.48 &  \textbf{32.44} & -  \\
\hc Sq-80\% & 0.225  & 16.66 & 24.70 & 41.88 & 21.10 & 15.91 & 9.13 & 27.00 & 24.68 & 24.23 & 70.00 & 60.81 & 39.37 & 42.07 & 41.89 & 32.31 & 32.82\\
\midrule
QUEST & 0.168 & 15.37 & 23.33 & 41.45 & 20.26 & 17.39 & 7.85 & 25.88 & 25.06 & 24.43 & 65.00 & 62.54 & - & 46.20 & 43.06 & 32.14 & -  \\
\hc Sq-90\% & 0.125 & 16.97 & 24.96 & 41.14 & 20.70 & 16.40 & 9.24 & 27.00 & 24.59 & 23.51 & 71.50 & 59.37 & 39.87 & 44.78 & 43.80 & \textbf{32.61} & 33.13 \\
\hd H-Sq-90\% & 0.118 & 16.69 & 24.79 & 40.38 & 20.78 & 16.21 & 8.91 & 25.02 & 24.77 & 22.34 & 70.50 & 58.23 & 39.40 & 44.31 & 43.34 & 32.02 & 32.55 \\
\midrule
\multicolumn{18}{c}{\textbf{LongChat-7B-v1.5-32K}}  \\ 
\midrule
All KV & 1  & 20.82 & 28.95 & 43.06 & 32.79 & 24.18 & 14.09 & 30.67 & 22.83 & 26.09 & 66.50 & 83.45 & 41.25 & 53.20 & 56.64 & 38.71 & 38.89\\
\midrule
\hc Sq-70\% & 0.325 & 20.93 & 29.18 & 43.00 & 33.02 & 23.61 & 14.55 & 31.13 & 22.93 & 26.25 & 66.50 & 83.60 & 40.90 & 54.64 & 56.93 & {38.94} & 39.08  \\
\midrule
QUEST & 0.215 & 19.33 & 31.51 & 41.65 & 31.79 & 23.25 & 12.58 & 31.09 & 22.84 & 26.87 & 67.50 & 84.33 & - & 53.57 & 55.37 &  38.59 &  -  \\
\hc Sq-80\% & 0.225 & 20.57 & 29.64 & 42.80 & 33.06 & 23.63 & 15.27 & 31.31 & 23.21 & 26.17 & 65.50 & 83.87 & 41.28 & 52.83 & 57.17 & \textbf{38.85} & 39.02 \\
\midrule
QUEST  & 0.168 & 18.03 & 30.21 & 37.83 & 31.78 & 21.03 & 11.21 & 30.52 & 22.84 & 26.47 & 63.50 & 84.71 & - & 51.50 & 55.82 & 37.34 &  -  \\
\hc Sq-90\% & 0.125 & 18.60 & 29.86 & 42.21 & 35.71 & 23.12 & 14.31 & 31.61 & 22.79 & 26.17 & 65.50 & 78.85 & 41.22 & 51.57 & 56.95 & \textbf{38.25} & 38.46 \\
\hd H-Sq-90\%  & 0.122 & 18.86 & 30.51 & 42.25 & 35.42 & 20.88 & 13.85 & 30.85 & 22.84 & 25.71 & 65.50 & 78.50 & 40.96 & 51.89 & 57.20 & 38.02 & 38.23 \\
\bottomrule
\end{tabular}
}
\vspace{-4.5mm}
\end{table*}

\section{Dataset for Fixed Context Processing}

\label{sec:bench}

Despite the growing demand for long-context applications where a fixed document is used to answer multiple user requests (e.g., code generation or long-document QA), 
there is currently no benchmark designed to test this scenario. 
Recent long context benchmarks \cite{bai2023longbench,hsieh2024ruler} do not evaluate the handling of multiple queries on the same document. 
This leads to a longer iteration cycle for developing fixed context optimization methods since the offline preprocessing step must be performed for every sample instead of once per fixed context.

To bridge this gap, we introduce \BENCH (Fixed-Prefix QA), a benchmark which evaluates the ability of LLMs to manage multiple queries on a single long-context document.
Our dataset curation pipeline (outlined in detail in Section \ref{sec:appendix-dataset}) involves collecting long documents from arXiv papers, and then collecting question-answer pairs based on each document using a multi-step generation and filtering procedure inspired by the Llama-3 training data generation approach~\cite{dubey2024llama}.
This process has yielded 1,127 high-quality question-answer pairs (24 on average per document) for our benchmark, ensuring a diverse and challenging benchmark for long-document question-answering.

\begin{figure*}[t]
\centering
\includegraphics[width=1\linewidth]{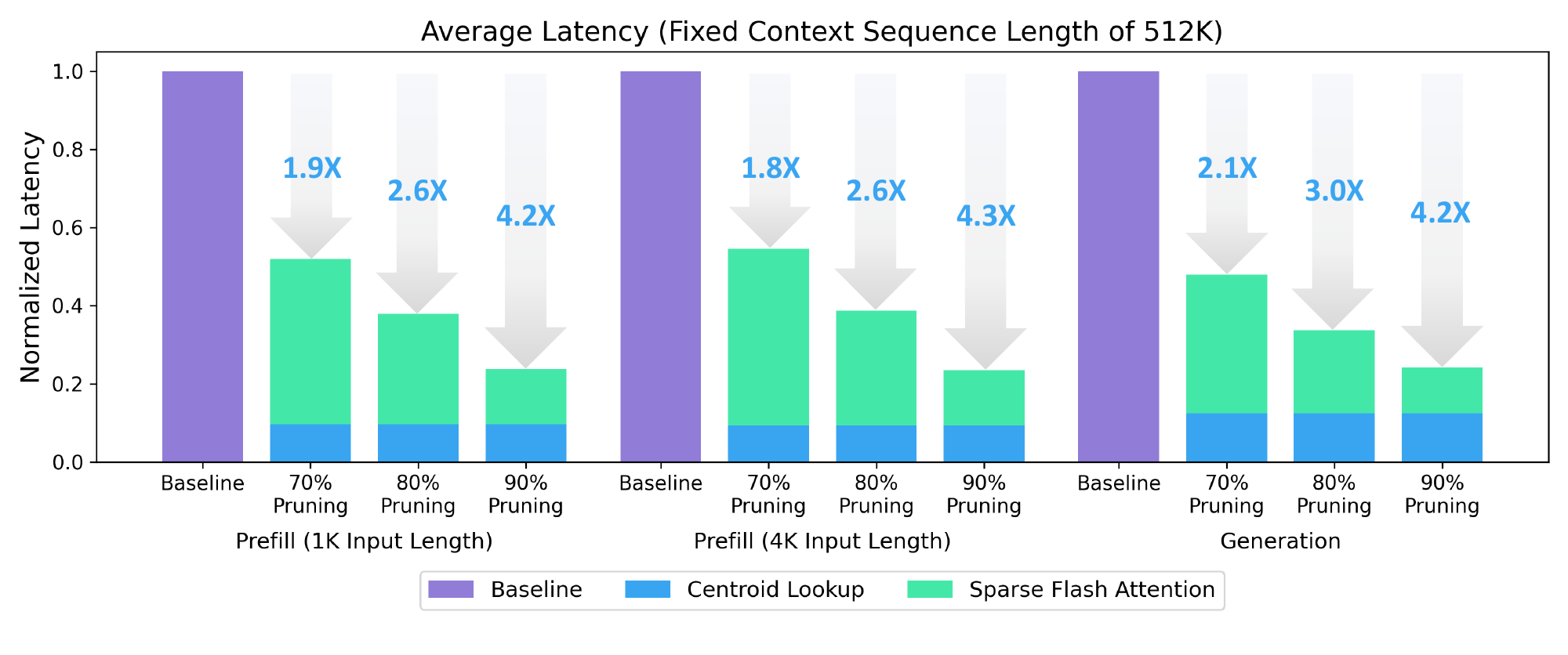}
\vspace{-5mm}
 \caption{
Kernel implementation latency results for FlashAttention baseline as well as for \OURS with 70\%, 80\%, and 90\% sparsity settings.
We report latency results for prefill (with 1K and 4K input length) as well as for generation with a single input token.
Latency results are normalized to the FlashAttention baseline runtime for prefill (and to our Triton FlashDecoding baseline for generation) for the same input length.
}
  \label{fig:latency}
\end{figure*}

\begin{table*}[h!]
\centering
\caption{
RULER evaluation results with \OURS. We report results across different RULER tasks for the LWM-Text-Chat-1M model using 32K context length for evaluation.
We also report the KV budget (``Budget''), which gives the expected percentage of the KV cache that needs to be loaded during inference for each configuration.
Our results show that our method is able to retain the accuracy of the baseline model, even with aggressive sparsity settings.
}
\label{tab:ruler}
\scriptsize
\setlength{\tabcolsep}{4.7pt}{

\begin{tabular}{c|c|ccccccccccccc|c}
\toprule
\textbf{Config} & 
\rotatebox{0}{\textbf{Budget}} & \rotatebox{0}{\textbf{Niah1}} & 
\rotatebox{0}{\textbf{Niah2}} & \rotatebox{0}{\textbf{Niah3}} & 
\rotatebox{0}{\textbf{MKey1}} & \rotatebox{0}{\textbf{MKey2}} & 
\rotatebox{0}{\textbf{MKey3}} & \rotatebox{0}{\textbf{MValue}} & 
\rotatebox{0}{\textbf{MQuery}} & \rotatebox{0}{\textbf{VT}} & 
\rotatebox{0}{\textbf{CWE}} & \rotatebox{0}{\textbf{FWE}} & 
\rotatebox{0}{\textbf{QA1}} & \rotatebox{0}{\textbf{QA2}}  & 
\rotatebox{0}{\textbf{Avg.}} \\ 
\midrule
All KV  &  1 & 100.0 & 100.0 & 99.4 & 100.0 & 99.6 & 96.4 & 45.6 & 35.3 & 58.4 & 9.7 & 66.7 & 63.2 & 43.6 & 70.6 \\
\hc Sq-70\% & 0.325 & 100.0 & 100.0 & 99.2 & 100.0 & 99.6 & 93.2 & 43.8 & 37.4 & 57.7 & 8.4 & 65.8 & 60.2 & 42.0 & 69.8 \\
\hd Sq-90\% & 0.125 & 100.0 & 100.0 & 98.6 & 100.0 & 99.4 & 81.8 & 42.1 & 36.0 & 51.4 & 9.7 & 65.1 & 57.8 & 41.6 & 68.0 \\
\bottomrule
\end{tabular}
\vspace{-3mm}
}
\end{table*}

\section{Results}
\label{sec:results}

\subsection{Experimental Details}

We evaluate our method on a range of downstream long context length tasks. 
We leverage the LongBench~\cite{bai2023longbench} and RULER~\cite{hsieh2024ruler} benchmark suites as well as PreFixQA for our evaluation. 
For our single-level experiments, we set the number of cluster centroids to be 5\% of the fixed context length, and for our hierarchical experiments we set the number of Level 1 centroids and Level 2 centroids to be 1\% and 5\% of the fixed context length, respectively. 
For our hierarchical experiments, we set the Level 1 threshold such that 50\% of the keys would be ruled out before performing the fine-grained Level 2 lookup.
For the single-level lookup, the metadata overhead was therefore 2.5\% of the KV cache memory footprint, since we only need to store centroids for the keys and not for the values.
For the hierarchical lookup, the metadata overhead was 3\% of the KV cache size in terms of storage requirements, but only approximately 1.75\% of the KV cache size in terms of metadata loaded during generation.
Additional experimental details are provided in Appendix \ref{appendix:experimental}.

\begin{table}[t]
\centering
\caption{PreFixQA evaluation results with \OURS. We report results using our single-level approach (``Sq'') and with our hierarchical lookup-based approach (``H-Sq''). LLaMA-2, LWM, and LongChat are LLaMA-2-7B-32K, LWM-Text-Chat-1M, and LongChat-7B-v1.5-32K, respectively.
}
\label{tab:prefixqaDataset}
\vspace{-2mm}
\scriptsize
\begin{tabular}{c*{3}{>{\centering\arraybackslash}p{1.4cm}}}
\toprule  
\textbf{Config} &
\rotatebox{0}{\textbf{{LLaMA-2}}} & 
\rotatebox{0}{\textbf{{LWM}}} & \rotatebox{0}{\textbf{{LongChat}}} \\
\midrule
All KV & 43.47 & 14.92 & 24.13 \\
\midrule
\hb Sq-70\% & 42.14 & 14.45 & 23.79 \\
\hc Sq-90\% & 36.95 & 14.25 & 23.94 \\
\hd H-Sq-90\% & 37.05 & 14.12 & 23.71 \\
\bottomrule
\end{tabular}
\vspace{-2mm}
\end{table}

\subsection{Accuracy Evaluation Results}
\label{subsec:eval}

We evaluate our method on long-context datasets from LongBench \cite{bai2023longbench}, a comprehensive benchmark which includes document QA, few-shot learning, and code completion tasks. 
Table \ref{tab:longbench} provides evaluation of our method on the non-synthetic English language tasks in LongBench for the LLaMA-2-7B-32K~\cite{llama-2-7b-32k}, LWM-Text-Chat-1M~\cite{liu2024world}, and Longchat-7B-v1.5-32K models \cite{longchat2023}. 
We also provide baseline comparisons with QUEST \cite{tang2024quest}.
To ensure a fair comparison, we set the token budget for their method dynamically for each input sample to match our approach.
We also provide Longbench evaluation for Llama-3.1-8B and Falcon3-7B in Appendix \ref{sec:appendix-llama3}.
Additional baseline configuration details are provided in Appendix \ref{appendix:experimental}.

The results show that our method provides similar accuracy to the full KV cache baseline on long context-length tasks, while offering significant efficiency improvement in terms of reduction in KV cache loading and attention computation.
Across all three models in Table \ref{tab:longbench}, our method maintains full KV cache accuracy with less than 0.11 point degradation at 70\% sparsity, reducing the KV budget by 3.1$\times$. 
Even at a more aggressive 90\% sparsity, which reduces the KV budget by 8$\times$, our method only introduces a small accuracy degradation of within 0.5 points.
Note that our method’s accuracy also matches an idealized baseline, where full attention is computed with all keys before retaining only the highest-scoring ones, as further discussed in Appendix \ref{appendix:ideal}.
This demonstrates that our method can effectively identify and retrieve the most relevant keys (i.e., those that yield high attention scores) without loading all the keys.

Furthermore, our method outperforms the QUEST baseline, with a pronounced accuracy gap of up to $\sim$1 point for more aggressive sparsity settings.
This highlights the advantage of semantic-based clustering for identifying and retaining important keys.
We also include results for our hierarchical lookup approach, demonstrating how our hierarchical method has lower overhead from the centroid lookup with minor accuracy loss relative to performing a single-level lookup.
Appendix \ref{appendix:cluster-granularity} provides an ablation comparing the accuracy of our algorithm when using coarse-grained versus fine-grained centroids, demonstrating how fine-grained centroids are required for high accuracy. 
Our hierarchical method can attain the high accuracy of fine-grained centroids with a reduced centroid lookup overhead.

Additionally, we present evaluation for our method on the \BENCH dataset in Table~\ref{tab:prefixqaDataset} using the LLaMA-2-7B-32K, Longchat-7B-v1.5-32K, and LWM-Text-Chat-1M models \cite{llama-2-7b-32k, longchat2023, liu2024world}.
The results demonstrate how our method provides similar accuracy as the baseline for fixed context use-cases, while significantly compressing the fixed context that needs to be dynamically loaded during inference. 
We also present an evaluation for our method on the RULER benchmark \cite{hsieh2024ruler} using the LWM-Text-Chat-1M model \cite{ liu2024world} in Table \ref{tab:ruler}, demonstrating the consistent performance for our method across different benchmark suites.

\subsection{System Evaluation}
\label{sec:system_results}

To evaluate our system implementation, we benchmarked the end-to-end runtime for our centroid lookup and sparse FlashAttention kernels (and include the runtime of Pytorch code for setting up arguments for our kernels), and compared these with baseline FlashAttention and FlashDecoding implementations.
We performed benchmarking on an NVIDIA H100 NVL GPU.
Experimental details for our kernel benchmarking experiments are provided in Appendix \ref{appendix:experimental}.
Figure \ref{fig:latency} shows the latency for the FlashAttention baseline, as well as for \OURS with 70\%, 80\%, and 90\% sparsity with 512K context length.
We set the number of centroids to be 5\% of the context length.
We report results for generation (one input token) as well as prefill with 1K and 4K input tokens, and we normalize the latency to the baseline latency for each input size.
These results show the benefits of our method for accelerating long context length inference, with 4.3$\times$ / 4.2$\times$ speedups demonstrated for the prefill and decode phases.
Additional results for context length 128K are provided in Appendix \ref{appendix:kernel-128k}. Hierarchical centroid lookup kernel benchmarking is also provided in Appendix \ref{appendix:hierarchical-kernel}, demonstrating how our hierarchical approach can accelerate the lookup runtime by up to 1.4$\times$ for 512K context length.
Appendix \ref{appendix:cluster-runtime} provides measured runtime for clustering the fixed context keys; note that this is performed \textit{offline} ahead of inference time, so it is not included in the inference latency breakdown.

\section{Conclusion}

We propose \OURS as a method for accelerating attention in long context-length applications.
Our method groups the fixed context keys by applying K-means clustering offline and leverages representative centroids for each cluster to quickly identify important keys.
Online during inference, we first compare the new input query with the representative centroids, and then only compute exact attention for these important keys.
Our method can be extended to the hierarchical retrieval scheme, which can reduce the memory and compute complexity of lookups to logarithmic complexity with respect to the fixed context length.
\OURS is able to provide 4.3$\times$ / 4.2$\times$ speedups during prefill and decode phases for long context inference, while maintaining accuracy.
Additionally, we outline how our algorithm can be extended using a hierarchical centroid lookup, allowing us to achieve the accuracy of fine-grained centroid lookups while maintaining the efficiency of coarse-grained centroids, thereby improving the scalability of our approach for longer context lengths.
Our approach accelerates long context length LLM inference with fixed context applications while maintaining accuracy.
\section{Limitations}

One of the limitations of our work is that both the sparsity threshold and the number of centroids used are hyperparameters.
Furthermore, the degree of sparsity that is attainable without accuracy degradation is also dependent on the input context and the type of task.
Our work could therefore be extended by developing an automated way of configuring these hyperparameters depending on the target accuracy level and the input context.
Our approach also focuses on accelerating fixed context applications, which limits its use for applications where the full context is only available online.
Future work can be done to accelerate the initial offline clustering step in order to allow our method to be used in online use-cases.
Finally, our current approach does not perform any approximation for the less important keys.
Future work could investigate methods to approximate the attention to these keys in order to compensate for this error.

\section{Acknowledgements}
We are grateful for the insightful discussions with Dhairya Malhotra.
We acknowledge gracious support from the FuriosaAI team including Jihoon Yoon, Suyeol Lee, and Hyung Il Koo, as well as from Intel, Apple, NVIDIA, and Mozilla.
We also appreciate the support from Microsoft through their Accelerating Foundation Model Research, including
great support from Sean Kuno.
Furthermore, we appreciate support from
Google Cloud, the Google TRC team, and specifically Jonathan Caton, and Prof. David Patterson.
Prof. Keutzer's lab is sponsored by the Intel corporation, Intel One-API, Intel VLAB team, the Intel One-API center of
excellence, as well as funding through BDD and BAIR.
We appreciate great feedback and support from Ellick Chan, Saurabh Tangri, Andres
Rodriguez, and Kittur Ganesh.
Sehoon Kim would like to acknowledge the support from the Korea Foundation for Advanced Studies (KFAS).
Michael W. Mahoney would also like to acknowledge
a J. P. Morgan Chase Faculty Research Award 
as well as 
the DOE, NSF, and ONR.
This work was supported by the Director, Office of Science, Office of Advanced Scientific Computing Research, of the U.S. Department of Energy under Contract No. DE-AC02-05CH11231.
Our conclusions do not necessarily reflect the position or the policy of our sponsors, and no official endorsement should be~inferred.

\bibliography{custom}

\appendix
\clearpage

\section{Related Work}

\label{sec:appendix-related-work}

This section provides a detailed discussion of related work for Long-Context LLMs as well as KV cache compression. 

\subsection{Long-Context LLMs}
With the growing popularity of long-context applications, there has been a continuous development of LLMs that can support context lengths exceeding 100k, and even up to 1M tokens. 
This includes proprietary models such as GPT-4-Turbo~\cite{achiam2023gpt}, Claude-2~\cite{claude2} and Gemini 1.5~\cite{gemini15}, which support context lengths of up to 128k, 200k, and 1M tokens, respectively.
On the open-source front, 
several efforts have been made to extend the context lengths beyond the length on which the original models were trained~\cite{longchat2023,chen2023longlora}.
A notable work is Large World Model (LWM)~\cite{liu2024world}, which has demonstrated extending the context length of Llama 2~\cite{touvron2023llama2} to 1M tokens. 
However, as context lengths increase, the KV cache often becomes a critical bottleneck, significantly impacting memory usage and latency during LLM inference~\cite{tang2024quest,hooper2024kvquant}.
Therefore, KV cache compression methods have emerged as a critical concern for enabling efficient inference when using long-context models.

\subsection{KV Cache Compression for Long-Context Inference}
To enable more efficient long-context inference by reducing the KV cache size, 
several methods have been proposed, including quantization~\cite{hooper2024kvquant,liu2024kivi,kang2024gear,liu2024intactkv}, shared KV cache across tokens~\cite{nawrot2024dynamic} and layers~\cite{brandon2024reducing}, 
and token pruning~\cite{fu2024lazyllm}. 
A notable approach which will be discussed in more detail is KV cache sparsification, which follows a prior line of work in attention sparsification~\cite{roy2021efficient,child2019generating,zaheer2020big}.
There are two general directions which have been pursued for KV cache sparsification: KV cache eviction, and sparsely loading the KV cache. 

\noindent
\textbf{KV Cache Eviction.}
KV eviction has become a widely used method for compressing the KV cache by identifying and removing less important tokens. 
Various strategies have been proposed to determine token importance, including attention score contribution~\cite{zhang2024h2o,oren2024transformers}, persistent attention patterns during generation~\cite{liu2024scissorhands}, token entropy~\cite{yao2024sirllm}, and additional heuristic-based policies~\cite{ge2023model}.

In use cases where long context prompts are followed by varying questions, 
the importance of the KV cache for the context should be decided on the basis of its relevance to the subsequent question.
To address this, SnapKV~\cite{li2024snapkv} proposes selecting KV cache entries solely based on the attention scores of the most recent prompt tokens to the rest of the input prompt. 
However, since the important tokens in the input prompt are determined once and remain fixed throughout the generation process, it cannot adapt to changing token importance during generation or in response to subsequent user inputs.
InfiniPot~\cite{kim2024infinipot} extends this idea by iteratively compressing the context based on its relevance to predefined task-specific prompts that resemble potential input questions.
Nevertheless, selecting important tokens offline using proxy prompts may not accurately reflect future queries.

Likewise, eviction-based approaches discard tokens and retain the remaining ones throughout generation, potentially overlooking the fact that discarded tokens could become important later in the process.
\OURS, on the other hand, bypasses the need for a full KV cache lookup by clustering the KV cache and retrieving only the most relevant clusters through an efficient centroid lookup. 
This approach is lightweight enough to be applied at every generation step, thereby ensuring relevant context is retrieved for every query token.

\noindent
\textbf{Sparse KV Cache Loading.}
One previous direction that has been explored aims to store the full KV cache, but only load in the relevant keys and values dynamically during inference. 
QUEST~\cite{tang2024quest} clusters consecutive KV cache entries and dynamically retrieves the most relevant clusters based on their relevance to each query token during generation. 
Another line of relevant work here is application of fast kernel summation methods~\cite{gray2000n,yang2003improved,lee2005dual,morariu2008automatic,march2015askit} and in particular variants of Fast Multipole Method (FMM)~\cite{coifman1993fast} which were originally proposed to accelerate N-body simulations.
In the context of Transformers, recent work of~\cite{kang2023fast} utilizes FMM to cluster consecutive past tokens and assign coarser-grained clusters to older tokens, reducing the memory overhead of storing the entire past tokens.
However, this approach, as well as QUEST~\cite{tang2024quest}, rely on \textit{physical} proximity for clustering, whereas in natural language applications clustering should instead be based on \textit{semantic} proximity, as tokens that are physically far apart can be semantically similar. 
\OURS addresses this by clustering tokens based on their embedding similarity, ensuring that semantically relevant tokens are retrieved for future generations.

Another prior line of work aims to leverage vector search methods for only loading important keys and values.
PQCache \cite{zhang2024pqcache} applied product quantization-based vector search to identify important keys.
RetrievalAttention \cite{liu2024retrievalattention} uses a K-Nearest Neighbors-based vector search approach, which offloads dynamic retrieval of important keys and values to the CPU.
However, these prior approaches are restricted to the generation stage and do not accelerate prefill, which is critical to reducing time-to-first-token (TTFT) latencies. 

In contrast with prior works which leverage vector search methods, \OURS uses a fast centroid lookup to enable accurate retrieval of relevant contexts on the GPU without requiring offloading operations to CPUs, as in~\cite{liu2024retrievalattention}.
Our approach is also able to accelerate both prefill and generation.
Furthermore, our method allows for loading more or fewer keys from different heads, depending on the number of important keys for each head.
This approach enables us to achieve higher accuracy while aggressively reducing the number of KV entries.

\subsection{Sparsity for Accelerating Prefill}

There has also been previous work on exploiting sparsity to accelerate prefill for LLM inference. 
One prior approach detected and exploited particular attention sparsity patterns in order to reduce the amount of computation required \cite{jiang2024minference}. 
A second prior approach is token pruning, where tokens are progressively dropped as we go from earlier layers to later layers in the network, thereby skipping computation for low-importance tokens at later layers \cite{yang2024pyramidinfer,fu2024lazyllm}.
Note that these approaches are mainly tailored for accelerating attention for long prefill use cases, whereas our method accelerates attention to the fixed context for both prefill with the user query and for decode.

\section{Additional Analysis} 
\label{appendix:analysis}

\subsection{t-SNE Visualization of Keys and Their Clusters}
\label{appendix:cluster}

\begin{figure*}[h]
  \centering
  \begin{subfigure}{0.43\textwidth} 
    \includegraphics[width=\linewidth]{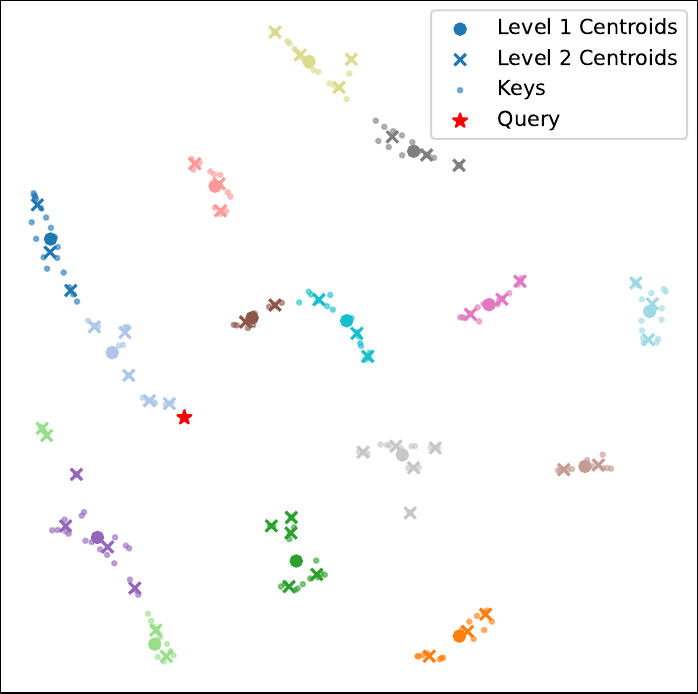}
    \label{fig:figure1}
  \end{subfigure}%
  \hspace{3mm}
  \begin{subfigure}{0.43\textwidth} 
    \includegraphics[width=\linewidth]{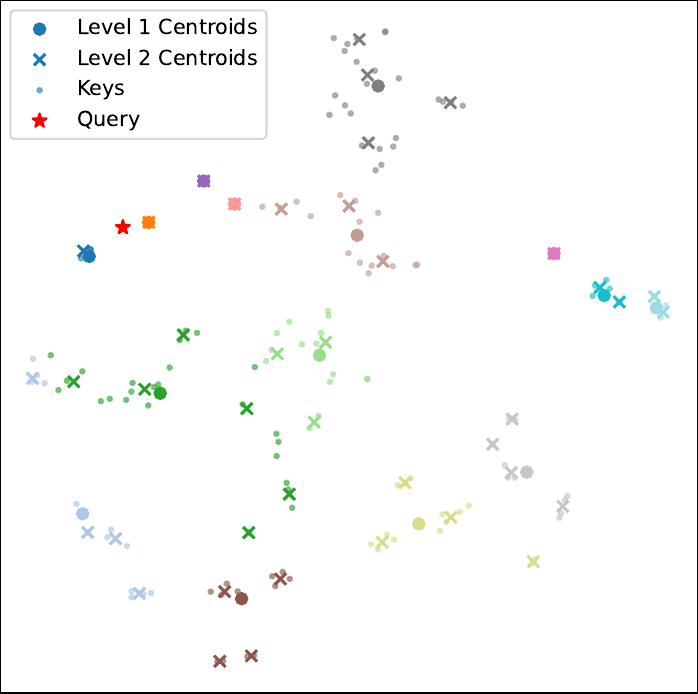}
    \label{fig:figure2}
  \end{subfigure}
  \vspace{-4mm}
  \caption{t-SNE visualization of key embeddings and their Level 1 and 2 clusters from LLaMA-2-7B-32K on the TREC benchmark (two attention heads, with index 24 and 25, from layer 0). For clarity, only the top 15 Level 1 clusters nearest to the query are shown.}
  \label{fig:tsne_key_clusters}
\end{figure*}

Figure~\ref{fig:tsne_key_clusters} illustrates t-SNE plots of key embeddings and their Level 1 and 2 clusters.
As can be seen, while the coarser Level 2 clusters offer a rough grouping of the keys, the finer Level 1 clusters allow for a more detailed and accurate representation within each cluster.

\subsection{Attention Score Skewness Analysis}
\label{appendix:skewness}

\begin{figure*}[h]
\centering
\includegraphics[width=\linewidth]{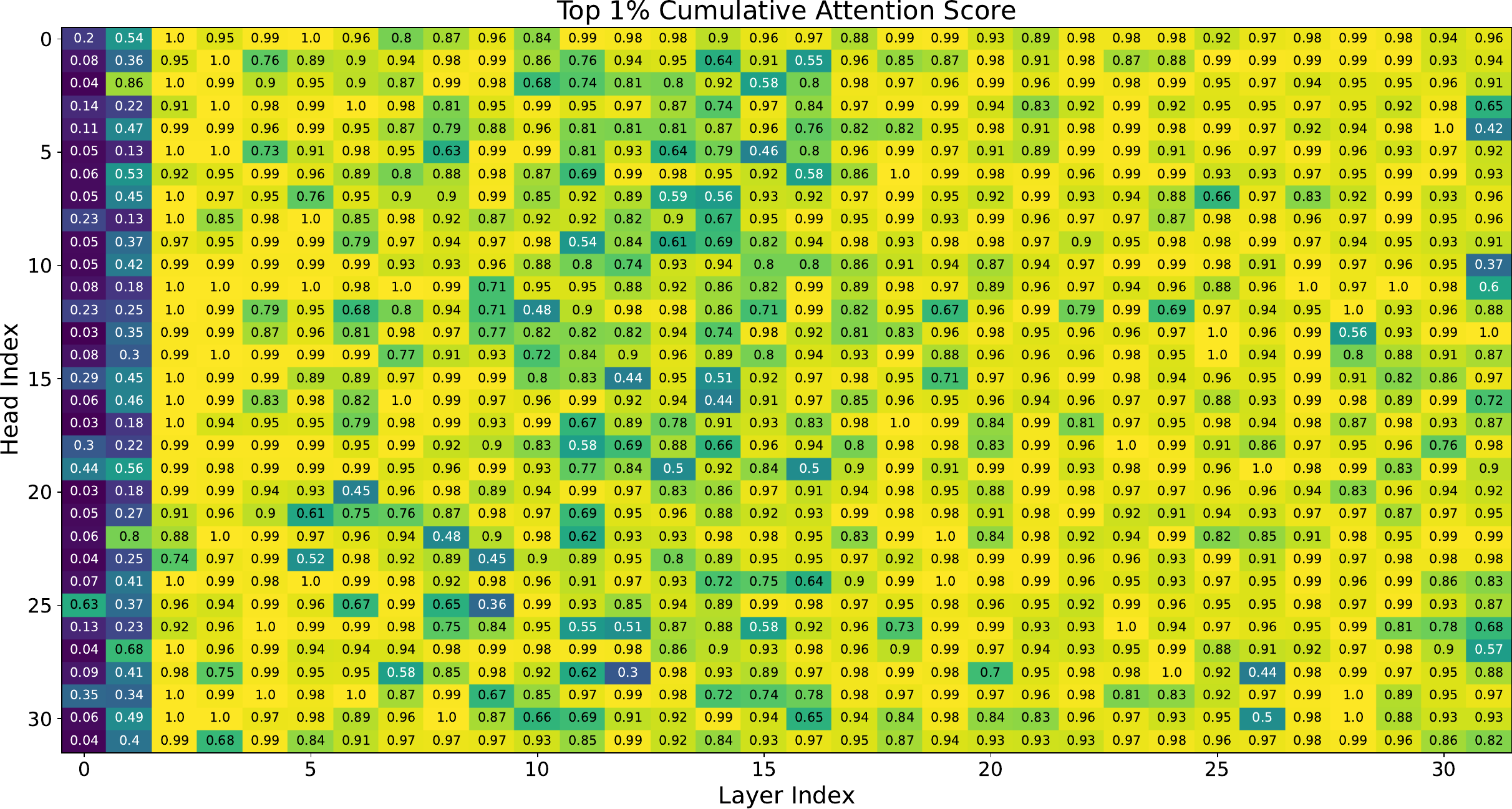}  
\caption{
Cumulative attention scores for the top 1\% highest scoring attention values across different heads and layers in LLaMA-2-7B-32K for a single sample on the TREC benchmark.}
  \label{fig:cumscore}
\end{figure*}

Figure~\ref{fig:cumscore} illustrates the cumulative attention scores for the top 1\% highest attention values within the same model. 
A higher value (up to a maximum of 1) indicates that the attention head has a sharper, more skewed distribution, while a lower value indicates a flatter distribution of attention scores.
This plot demonstrates how the attention heads in the first two layers of the LLaMA-2-7B-32K model, as well as a subset of the heads at each of the remaining layers, have a flatter distribution of attention scores, and therefore we need to load in more keys for accurate computation of attention.

\section{Global Threshold Calibration}
\label{appendix:calibration}
To determine the appropriate global threshold for achieving a target sparsity level, we use a small number of tokens (100 tokens) from the end of the fixed context in our experiments to calibrate the threshold offline. 
We compute the average $S_i$ scores using these last few query tokens and the remaining fixed context keys, and set a threshold $T$ that achieves the desired sparsity level.
Note that we exclude the last keys during calibration 
to avoid the impact of causal masking; these tokens are thus retained exactly and not clustered. 
Once the global threshold is decided, it is kept the same throughout the prefill and generation stages.

\section{Centroid Lookup System Implementation Details}

\label{appendix:centroid-lookup}

\indent
\textbf{Prefill Stage.} 
During prefill, where multiple query tokens are available, we split the workload along the query sequence length dimension as in FlashAttention-2 \cite{dao2023flashattention} to attain additional parallelism.
Since this process produces individual $S_i$ values for each query token, 
we compute their average to obtain $\bar{S_i}$, i.e., the \textit{averaged} importance score for each key cluster across all query tokens.
We then check whether $\bar{S_i} > T$ to determine whether to load the keys in the corresponding cluster.
Since $S_i$ is an estimate of the Softmax value, which is normalized to sum to 1, averaging across query tokens provides a simple way to calculate their combined importance score.

\noindent
\textbf{Generation Stage.}
During generation, achieving parallelism is more challenging, as we cannot leverage parallelism across the dimension of the length of the query sequence. 
This is particularly problematic when dealing with small batch sizes, as in that case the only parallelism we can leverage is across different heads.
To accelerate centroid lookup during generation, we additionally compute and store $\exp{(q C_{i}^\top)}$ for each cluster during the first pass over the key centroids while we are computing the denominator $D = \sum_j N_j \cdot \exp{(q C_{j}^\top)}$.
Then, in the second pass, we load the precomputed $\exp{(q C_{i}^\top)}$ values and compare them against $DT$ to determine the importance of each cluster, without the need to explicitly compute $S_i$.
This second pass can be parallelized across the cluster dimension for fast comparison.
As highlighted in Appendix \ref{appendix:kernel-ablation}, these optimizations are crucial for performing the centroid lookup during generation without substantial latency overhead.
Similar to the numerically stable implementation of Softmax, where the maximum value is subtracted from all inputs, our centroid lookup approach during the generation stage also subtracts the maximum value from all inputs while computing the denominator $D$.
Then, when comparing $\exp{(q C_{i}^\top)}$ to the threshold $DT$, the maximum value correction can similarly be accounted for by scaling the threshold $DT$ using the exponential of the maximum value rather than by scaling $\exp{(q C_{i}^\top)}$ down by this value. 

\noindent
\textbf{Hierarchical Centroid Lookup Kernel.}
We also implement hierarchical centroid lookup kernels for generation (see Appendix \ref{appendix:hierarchical-kernel} for benchmarking experiments). 
The comparison with the level one centroids is identical to the one-level lookup kernel for generation, and it outputs the indices of the level two centroids which we need to compare with.
The kernel for the second or later levels of the hierarchy takes as input the indices of the centroids which need to be loaded.
This kernel then performs similar parallelization as the generation kernel in the one-level case, except that when we parallelize the work across the cluster dimension we ensure that a balanced number of centroids are mapped to each GPU SM.
This is analogous to how we handle unbalanced sparsity in the sparse FlashAttention kernel, as described in Section \ref{subsec:sparse-flash-attn}.

\section{Overview of Dataset Generation Pipeline}

\label{sec:appendix-dataset}

This section discusses in detail the dataset curation pipeline for \BENCH (Fixed-Prefix QA), which evaluates the ability of LLMs to manage multiple queries on a single long-context document.
Our dataset curation pipeline consists of two phases: (i) collecting long documents from arXiv papers; and (ii) generating question-answer pairs based on each document, while ensuring correctness and consistency.
We also provide the GPT-4-Turbo~\cite{achiam2023gpt} prompts that were used to generate and filter high-quality question-answer pairs.

\subsection{Long Documents Collection.}
\BENCH is a long-document QA benchmark designed for one-document-multi-user-input scenarios. 
To collect high-quality long documents, we have sampled 47 papers from arXiv, each ranging from 17,000 to 200,000 characters, with an average length of 20 pages each after deleting references and appendices. 
To evaluate LLMs' capability to understand diverse types of content, 
we have selected papers from various fields, including computer science, electrical engineering, biology, machine learning, economics, and finance. 
To prevent training set contamination, all papers were sourced from 2024 using the arXiv API~\cite{arxivapi}.

\subsection{Question and Answer Generation.}
To generate multiple questions per each document, we have implemented a multi-step generation and filtering procedure inspired by the Llama-3 training data generation approach~\cite{dubey2024llama}.
To ensure that questions cover different sections of the document and avoid redundant questions by focusing too heavily on one part,
we divide each document into multiple chunks.
Each chunk is then provided to GPT-4-Turbo~\cite{achiam2023gpt} to generate potential questions that can be answered in 1-2 words.
We generate questions with short answers to enable more accurate evaluation through string comparison.

However, a single pass of question generation often results in low-quality question-answer pairs due to incorrectness or inconsistency of the answers. 
To avoid this, we introduce an additional filtering process to ensure the correctness and consistency of each question-answer pair.
In this step, each question (along with the specific chunk) is provided to GPT-4-Turbo five separate times to produce potential answers. 
We then filter out questions with inconsistent answers, which typically arise from ambiguity in the question or context. 
Furthermore, GPT-4-Turbo is used as a judge to score the correctness of each answer, based on the full document, on a scale of zero to ten. 
Questions where at least three out of five answers score above eight are kept; otherwise, they are discarded. For the retained questions, the highest-rated answer is selected and kept in the dataset.

\subsection{Prompts}

\textbf{Dataset curation prompts for question generation.} 
\begin{boxA}
    \texttt{\textbf{System:} You are a helpful assistant tasked with generating very specific, unambiguous, short-answer questions based on a provided Document. The goal is to create high-quality synthetic data. Ensure the following:} \\
        \texttt{1. The question must be fact-based and directly answerable from the text of the provided Document and section of the paper.}\\
        \texttt{2. The question should not be vague, subjective, or open to interpretation.}\\
        \texttt{3. Focus on creating concise, precise questions that yield a 1-2 word answer.} \\
        \texttt{4. Questions should be relevant to the section of the paper and avoid overly broad phrasing.} \\
        \texttt{5. Avoid generating questions where the answer is too complex or requires long explanations.} \\\\
        \texttt{\textbf{User:} Based on the section of the paper from the given document which is an arxiv paper generate one short-answer question that asks for specific information retrievable directly from the section of the paper. The answer must be 1-2 words only.} \\\\
        \texttt{Follow the format of this example:}\\
        \texttt{Example: Question: What type of buffer is integrated with warp-specialization in FlashAttention-3? Use this section of the paper as the context: "Given section". Do not output the answer; Just the question.}
\end{boxA}

\textbf{Dataset curation prompts for answer generation.} 
\begin{boxA}
    \texttt{\textbf{System:} You are a helpful assistant designed to generate accurate and specific short answers from a provided section of the paper and Document. Ensure the following: }
    \\
    \texttt{1. The answer must be concise (1-2 words).} \\
    \texttt{2. The answer must be directly retrieved from the provided text.}\\
    \texttt{3. If the section of the paper does not contain the information necessary to answer the question, respond with: 'The document does not contain the answer to the question.'.}\\
    \texttt{4. Avoid providing additional commentary, and only output the answer.}\\\\
   \texttt{\textbf{User:} Given the section of the paper below from an arXiv paper, generate a concise (1-2 words) answer to the following question. Retrieve the answer from the paper and the provided paragraph. Pay close attention to the document and retrieve the right answer. Output the answer only and not the question or anything else.}\\\\
   \texttt{Follow the format of this example:}\\
    \texttt{Example: Question: What type of buffer is integrated with warp-specialization in FlashAttention-3? Answer: circular SMEM buffer. Here is the section of the paper: "Given Section". Question: "Question". Answer:}
\end{boxA}

\textbf{Dataset curation prompts for filtering.}  
\begin{boxA}
    \texttt{\textbf{System:} Please act as an impartial judge and evaluate the quality of the question and answer pairs provided. You will be assessing them based on the provided document. Your evaluation should emphasize the following criteria:}\\
        \texttt{1. Correctness: Is the answer factually accurate and grounded in the document?} \\
        \texttt{2. Agreement: Does the answer directly address the question and provide a relevant response?} \\
        \texttt{3. Confidence: Does the answer confidently engage with the question, even if the Document does not contain the exact information?} \\\\
    \texttt{Important considerations for rating:}\\
         \texttt{ - Rate vague or overly general questions lower, especially if they lack specificity or do not make sense in the context of the document.}\\
         \texttt{ - Rate answers where the model, dataset, or method is unclear or missing details lower.}\\
         \texttt{ - If the answer states that the information is not in the document, confirm by reviewing the document. If the information is indeed missing, rate the answer highly. If it is present, rate the answer lower.}\\
         \texttt{ - Avoid focusing on questions about appendix numbers, or formatting details (like section names).}\\
         \texttt{ - Avoid asking questions that have 2 possible answers. if there are 2 possible answers and only one is provided, rate the answers low.}\\
         \texttt{ - If there are questions that are taken from the 'References' and 'Acknowledgments' sections, rate the answers low.}\\
    \texttt{For each answer, provide a rating on a scale from 1 to 10 based on its quality. }\\\\ 
    \texttt{\textbf{User:} The question is "Question" and the answers are "Answers". Make sure to give the rating for each answer in the answers list. Please output your rating in the following format:}\\ 
    \texttt{Question: "Question" Answer: [Answer1]- Rating: [[5]] Answer: [Answer2]- Rating: [[8]] ...}
\end{boxA}

\section{Experimental Details}
\label{appendix:experimental}

\noindent
\textbf{Evaluation.} Across all tasks, when identifying the ``fixed context'' portion of the input, we isolate the context before the user input using the prompt template for each task, and then we apply our approach to this fixed portion of the prompt.
For measuring accuracy with PreFixQA, we use F1 Score to calculate the similarity score between the outputs and the ground truth (as is used in LongBench for single-document QA tasks \cite{bai2023longbench}).
We use 32K as the maximum context length throughout our evaluation, and we truncate longer inputs for both LongBench and~PreFixQA. 
For RULER \cite{hsieh2024ruler}, we use the default configuration with 500 samples to evaluate our method.

\noindent
\textbf{KV Budget Computation.}
We report KV budget estimates throughout the evaluation based on the configured sparsity threshold and percentage of centroids used.
Note that the KV budget does not include performing recomputation with the same key centroid (as our current kernel implementation for prefill loads the key centroid twice to avoid materializing intermediate tensors).
{
Additionally, due to our calibration procedure (which sets a single threshold for both prefill and generation), the KV cache budget may be slightly higher than expected during prefill, and slightly lower than expected during generation.
This occurs since averaging the attention to the centroids across query tokens flattens the attention distribution, which leads to preserving more key tokens during prefill.
}
Also, with the hierarchical method, the portion of KV tokens that are loaded may deviate further from the expected value from calibration due to the potential for incorrectly filtering out important keys when comparing with the Level 1 centroids, and due to not loading all of the Level 2 keys when computing the denominator in Equation \ref{eq:centroid_lookup_hier_2}.
For our hierarchical lookup experiments, we therefore profiled the KV cache budget estimates reported in our evaluation.

\noindent
\textbf{Baseline Methods.} QUEST \cite{tang2024quest} uses a fixed token budget across all input samples when evaluating on LongBench.
In order to perform a fair comparison with our method, we set the token budget for their approach to be a fixed percentage of the length of each sample.
We dynamically set their token budget to be a percentage of the fixed context length (rounded up to the nearest multiple of 16).
Since our method only approximates attention to the fixed-context portion of the input, we adjust this dynamically computed token budget using the full input length for the sample as well as the maximum generation length for the target task. 
This adjustment ensures that the achieved compression ratio for the fixed context with their method is comparable with our approach.
Note that this adjustment also accounts for the 100 tokens at the end of the fixed context that we use for calibration purposes and retain exactly with our method.
For QUEST comparisons, we also leave the first two layers uncompressed to match their default configuration.
We use 90\% and 95\% sparsity settings for their method to obtain the two configurations reported in Table \ref{tab:longbench}, and we use 85\% sparsity settings for their method to obtain the configurations reported in Table \ref{tab:llama3}.

\noindent
\textbf{Kernel Benchmarking.} In order to benchmark our kernel implementations for long context length inference, we used sample text from the PG-19 language modeling dataset \cite{rae1911compressive}, and applied our clustering method to derive centroids to use when benchmarking.
We used PG-19 data since language modeling data allowed us to segment the fixed context and input into the desired lengths for measuring latency.
Using real data ensured that we had a realistic sparsity distribution across different heads and layers.
We ran clustering offline using offloading to collect data with a context length of 512K, and loaded this in one layer at a time in order to collect measurements.
We report the average runtime across all layers in all our experimental results.

For prefill, we benchmarked our Triton kernel implementations using \texttt{triton.testing.do\_bench} with 100 warmup runs and 500 measurement runs.
For generation, we used 50 measurement runs for 100 different input query tokens, and averaged the runtime across all of these runs. 
We use the H100 NVL hardware platform for our experiments.
We benchmarked the end-to-end runtime for our centroid lookup and sparse FlashAttention kernels (and include the runtime of Pytorch code for setting up arguments for our kernels).
For prefill, we compared the performance of our implementation with the FlashAttention-2 implementation provided through the PyTorch \texttt{scaled\_dot\_product\_attention} API.
For generation, we implemented a Triton FlashDecoding kernel optimized for single-batch inference to serve as a stronger baseline.

\section{Additional Longbench Evaluation}
\label{sec:appendix-llama3}

Table \ref{tab:llama3} provides evaluation of our method for the Llama-3.1-8B \cite{grattafiori2024llama} and Falcon3-7B \cite{Falcon3} models, as well as a comparison with QUEST \cite{tang2024quest} (with sparsity level configured to match our approach).
Note that these models use grouped query attention, which means that the keys and values are shared amongst a subset of the heads.
We allowed each head to independently select important keys and values (since this aligns with the open-source QUEST implementation), and then reported the total profiled memory footprint.
Future work could explore better aggregation strategies across query heads that share keys and values (eg. using the average score across query heads to allow each head to ``vote'' for the keys and values to retain); however, we let each head separately identify important keys and values for fair comparison with QUEST.
While we observe more substantial average accuracy degradation for Llama-3.1-8B relative to the uncompressed baseline (when compared with the degradation reported with our method for LLaMA-2-7B-32K, LWM-Text-Chat-1M, and Longchat-7B-v1.5-32K models in Table \ref{tab:longbench}), our method retains significantly higher accuracy than QUEST for the same compression ratio with both models.

\begin{table*}[h!]
\centering
\caption{
Longbench evaluation for \OURS with the Llama-3.1-8B and Falcon3-7B models, including the average score across tasks.
We also report baseline comparisons with QUEST \cite{tang2024quest}. 
We report the average score across LongBench tasks, as well as the average without SamSum (``All*'') for comparison against QUEST, whose evaluation framework does not support SamSum.
We also include the KV budget  for our method (``Budget''), which gives the percentage of the KV cache that needs to be loaded in during inference (including extra memory movement for cluster centroids).
We profiled the KV budget since different query heads which share the KV cache may require loading different keys and values.
We demonstrate how our method attains significantly closer accuracy to the baseline relative to QUEST for the same memory budget.
}
\label{tab:llama3}
\scriptsize
\setlength{\tabcolsep}{3.2pt}{
\begin{tabular}{c|c|ccc|ccc|ccc|ccc|cc|cc}
\toprule
&  \multirow{2}{*}{\raisebox{-4.8mm}{\textbf{Budget}}} & \multicolumn{3}{c|}{\textbf{Single-Doc. QA}} & \multicolumn{3}{c|}{\textbf{Multi-Doc. QA}} & \multicolumn{3}{c|}{\textbf{Summarization}} & \multicolumn{3}{c|}{\textbf{Few-shot Learning}} & \multicolumn{2}{c|}{\textbf{Code}} & \multicolumn{2}{c}{\textbf{Avg.}} \\
\cline{3-18}
\raisebox{4.2mm}{\textbf{Config}} & & \rotatebox{45}{\textbf{NQA}} & \rotatebox{45}{\textbf{Qspr}} & \rotatebox{45}{\textbf{MFQA}} & \rotatebox{45}{\textbf{HPQA}} & \rotatebox{45}{\textbf{2Wiki}} & \rotatebox{45}{\textbf{Musique}} & \rotatebox{45}{\textbf{GRep}} & \rotatebox{45}{\textbf{QMSum  }} & \rotatebox{45}{\textbf{MNews}} & \rotatebox{45}{\textbf{TREC}} & \rotatebox{45}{\textbf{TQA}} & \rotatebox{45}{\textbf{SSum}}  & \rotatebox{45}{\textbf{RB}}& \rotatebox{45}{\textbf{LCC}} & \raisebox{2.4mm}{\textbf{All*}}  & \raisebox{2.4mm}{\textbf{All}} \\ 
\midrule
\multicolumn{18}{c}{\textbf{Llama-3.1-8B}}  \\
\midrule
All KV & 1  & 21.91 & 12.62 & 33.90 & 11.64 & 13.90 & 8.62 & 30.07 & 25.46 & 2.38 & 73.5 & 90.97 & 47.43 & 70.39 & 67.83 & 35.63 & 36.47\\
\midrule
QUEST & 0.375 & 16.97 & 12.04 & 28.84 & 10.60 & 11.80 & 7.42 & 28.50 & 22.98 & 3.07 & 70.00 & 90.13 & - & 66.63 & 63.83 & 33.29 & - \\
\hd Sq-80\% & 0.360 & 19.13 & 12.50 & 32.43 & 11.51 & 12.57 & 8.06 & 28.44 & 23.85 & 5.93 & 72.50 & 90.73 & 45.05 & 65.55 & 65.20 & \textbf{34.49} & 35.25\\
\midrule
\multicolumn{18}{c}{\textbf{Falcon3-7B}}  \\ 
\midrule
All KV & 1  & 25.86 & 12.23 & 33.61 & 10.74 & 12.33 & 6.94 & 33.40 & 23.21 & 25.12 & 77.50 & 89.38 & 43.74 & 61.61 & 68.16 & 36.93 & 37.42 \\
\midrule
QUEST & 0.423 & 19.97 & 10.08 & 32.00 & 10.92 & 12.29 & 5.05 & 28.65 & 23.09 & 23.63 & 75.00 & 89.73 & - & 60.32 & 61.75 & 34.81 &  -  \\
\hd Sq-80\% & 0.302 & 24.37 & 12.83 & 33.15 & 11.58 & 12.23 & 5.48 & 32.65 & 23.47 & 25.23 & 75.00 & 88.87 & 40.88 & 61.93 & 67.84 & \textbf{36.51} & 36.82  \\
\bottomrule
\end{tabular}
}
\end{table*}

\section{Comparison with Ideal Lookup}

\label{appendix:ideal}

Table \ref{tab:ideal} provides comparisons with a baseline using an idealized lookup.
For the ``Ideal'' baseline comparisons, we first compute attention from the user input query tokens to all of the fixed context keys.
We then select the keys whose attention scores are above the configured threshold, and compute exact attention using only these keys.
This serves as an upper bound on the attainable accuracy for a given sparsity percentage, since it lets us identify the high-scoring tokens exactly before computing attention using these tokens. 
For this idealized baseline, we also calibrate for a global threshold across all layers to allow this approach to adaptively retain more or fewer keys for different heads, which allows us to make a fair comparison between our method and this idealized baseline.

\begin{table*}[h!]
\centering
\caption{
Ablation showing the accuracy of \OURS compared with an idealized baseline.
We report LongBench evaluation results for the Llama-2-7B-32K model, including the average score across tasks.
We also include the KV budget  for our method (``Budget''), which gives the percentage of the KV cache that needs to be loaded in during inference (including extra memory movement for cluster centroids).
Our results demonstrate that our centroid lookup method attains similar accuracy to the idealized baseline for the same level of sparsity across different LongBench tasks.
}
\label{tab:ideal}
\scriptsize
\setlength{\tabcolsep}{3.2pt}{
\begin{tabular}{c|c|ccc|ccc|ccc|ccc|cc|c}
\toprule
&  \multirow{2}{*}{\raisebox{-4.8mm}{\textbf{Budget}}} & \multicolumn{3}{c|}{\textbf{Single-Document QA}} & \multicolumn{3}{c|}{\textbf{Multi-Document QA}} & \multicolumn{3}{c|}{\textbf{Summarization}} & \multicolumn{3}{c|}{\textbf{Few-shot Learning}} & \multicolumn{2}{c|}{\textbf{Code}} & \multirow{2}{*}{\raisebox{-4.8mm}{\textbf{Avg.}}} \\
\cline{3-17}

\raisebox{4.2mm}{\textbf{Config}} & & \rotatebox{45}{\textbf{NQA}} & \rotatebox{45}{\textbf{Qspr}} & \rotatebox{45}{\textbf{MFQA}} & \rotatebox{45}{\textbf{HPQA}} & \rotatebox{45}{\textbf{2Wiki}} & \rotatebox{45}{\textbf{Musique}} & \rotatebox{45}{\textbf{GRep}} & \rotatebox{45}{\textbf{QMSum  }} & \rotatebox{45}{\textbf{MNews}} & \rotatebox{45}{\textbf{TREC}} & \rotatebox{45}{\textbf{TQA}} & \rotatebox{45}{\textbf{SSum}}  & \rotatebox{45}{\textbf{RB}}& \rotatebox{45}{\textbf{LCC}} &  \\ 
\midrule
All KV &  1 & 17.91 & 11.12 & 33.87 & 12.45 & 11.95 & 6.54 & 29.37 & 16.93 & 21.58 & 71.50 & 87.96 & 43.87 & 61.45 & 59.14 & 34.69  \\
\midrule
Ideal-70\% & 0.3 & 18.28 & 11.25 & 34.13 & 12.56 & 12.13 & 6.57 & 29.12 & 16.99 & 21.37 & 70.00 & 87.79 & 43.59 & 62.01 & 59.36 & 34.65 \\
\hd Sq-70\% & 0.325  & 18.55 & 11.78 & 34.33 & 12.31 & 12.31 & 6.26 & 29.50 & 16.90 & 20.76 & 69.00 & 87.96 & 43.90 & 61.29 & 59.53 & 34.60 \\
\midrule
Ideal-90\% & 0.1 & 17.65 & 11.77 & 33.89 & 12.67 & 11.86 & 6.04 & 29.02 & 16.79 & 23.22 & 69.00 & 87.23 & 44.33 & 57.82 & 60.17 & 34.39\\
\hd Sq-90\%  & 0.125 & 18.15 & 14.39 & 32.38 & 11.84 & 11.70 & 6.45 & 29.06 & 16.93 & 21.66 & 70.00 & 87.43 & 45.15 & 58.79 & 59.37 &  34.52 \\
\bottomrule
\end{tabular}
}
\end{table*}

\section{Clustering Granularity Ablation}
\label{appendix:cluster-granularity}

Table \ref{tab:cluster-granularity} includes empirical results showing the average accuracy on a subset of LongBench for the LWM-Text-Chat-1M model with different numbers of centroids and pruning percentages to analyze the impact of cluster granularity.
We find that with 5\% centroids (one centroid per 20 tokens), we attain high accuracy for aggressive (90\%) pruning.
With 1\% centroids (one centroid per 100 tokens), we observe large accuracy penalties with 90\% pruning; however, the centroid lookup overhead is reduced substantially.
This motivates our hierarchical method, which allows us to achieve the high accuracy of the lookup with fine-grained centroids (5\%) while maintaining the low lookup overhead with coarse-grained centroids (1\%).
The reduced lookup overhead is attained by first performing a coarse-grained lookup to filter out less promising fine-grained centroids, before comparing with only the more promising fine-grained centroids.

\begin{table}[t]
\centering
\caption{Average accuracy on a subset of LongBench tasks (TREC, 2WikiMQA, and MultifieldQA) for \OURS with the LWM-Text-Chat-1M model, with different numbers of centroids and pruning percentages.
With coarse-grained centroids (1\%, or one centroid per 100 tokens), we observe substantial accuracy degradation for aggressive sparsity (90\% pruning), whereas with fine-grained centroids (5\%, or one centroid per 20 tokens), we can attain high accuracy even with aggressive sparsity.
}
\label{tab:cluster-granularity}
\scriptsize
\begin{tabular}{cccc}
\toprule  
\textbf{Config} &
\textbf{1\% Centroids} & 
\textbf{2.5\% Centroids} & 
\textbf{5\% Centroids} \\
\midrule
All KV & 43.07 & 43.07 & 43.07 \\
\midrule
50\% Pruning & 40.72 & 42.93 & 42.87 \\
90\% Pruning &  19.55 & 42.19 & 42.97 \\
\bottomrule
\end{tabular}
\end{table}

\section{Kernel Benchmarking for 128K Sequence Length}

\label{appendix:kernel-128k}

\begin{figure*}[h]
\centering
\includegraphics[width=0.95\linewidth]{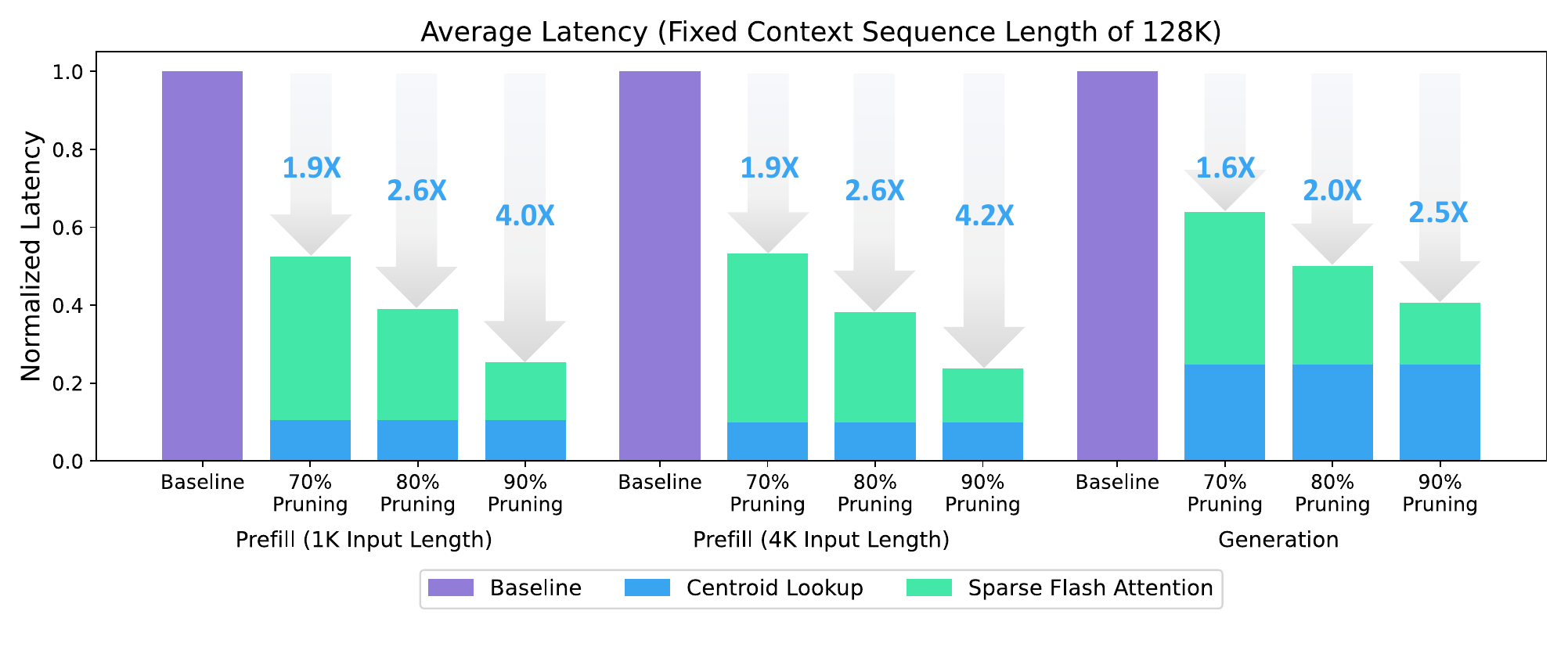}
 \caption{
Kernel implementation latency results for FlashAttention baseline as well as for \OURS with 70\%, 80\%, and 90\% sparsity settings.
We report latency results for prefill (with 1K and 4K input length) as well as for generation with a single input token.
Latency results are normalized to the FlashAttention baseline runtime for prefill (and to our Triton FlashDecoding baseline for generation) for the same input length.
}
  \label{fig:latency-128k}
\end{figure*}

Figure \ref{fig:latency-128k} shows the latency for the FlashAttention baseline as well as for \OURS with 70\%, 80\%, and 90\% sparsity with 128k context length, with the number of centroids set to be 5\% of the context length.
We report results for generation as well as prefill with 1K and 4K input tokens, and we report the latency for each configuration normalized to the baseline latency for the corresponding input size.
For prefill, we observe 4.2$\times$ speedups with 90\% sparsity, which is comparable with our speedups reported for 512K.
For generation, we observe reduced speedups, observing 2.5$\times$ speedups with 90\% sparsity, relative to the FlashDecoding baseline implementation.
The reduced speedups in this regime are due to greater overheads with the centroid lookup kernel for shorter sequence lengths.

\section{Generation Centroid Lookup Kernel Ablation}
\label{appendix:kernel-ablation}

We provide ablations for our centroid lookup kernel implementation for generation.
Specifically, we ablate the benefits of our single-pass optimization, as well as the improvements from parallelizing along the KV sequence length dimension in order to accelerate the centroid lookup during generation.
Table \ref{tab:kernel-ablation} shows the results for this ablation, demonstrating how these optimizations allow \OURS to achieve greater speedups during generation.

\begin{table*}[h!]
\centering
\caption{
Ablation for our centroid lookup kernel implementation during generation with sequence length 512K, showing the normalized latency relative to the baseline Triton FlashDecoding kernel.
We show the benefits of incorporating our single-pass optimization, as well as the gains from splitting along the KV dimension as in FlashDecoding \cite{flashdecoding}. 
These results highlight the importance of our lookup kernel optimizations for attaining speedups with \OURS during generation.
}
\vspace{2mm}
\label{tab:kernel-ablation}
\scriptsize
\begin{tabular}{c|c|ccc}
\toprule
Configuration &  FlashDecoding Baseline & Centroid Lookup & + Single-Pass Optimization & + Split-KV Optimization  \\
\midrule
Normalized Latency &  1  & 0.29 & 0.22 & 0.12\\

\bottomrule
\end{tabular}
\end{table*}

\section{Hierarchical Centroid Lookup Kernel Benchmarking}
\label{appendix:hierarchical-kernel}

We provide additional experimental results comparing the latency improvements that are attainable through leveraging our hierarchical centroid lookup kernel implementation. 
Table \ref{tab:hierarchical-kernel} provides kernel benchmarking experiments for our hierarchical lookup, demonstrating how our hierarchical centroid lookup can reduce the overhead of the single-level centroid lookup by up to 1.4$\times$ for a context length of 512K, which is particularly valuable for more aggressive sparsity thresholds where the cost of the centroid lookup is a greater contributor to overall latency.

\begin{table*}[h!]
\centering
\caption{
Latency for hierarchical centroid lookup during generation with a context length of 512K (normalized to the latency of the one-level lookup). The baseline latency is computed using one-level lookup with 5\% centroids. Reported latency results assume 90\% sparsity threshold for the level one lookup and >90\% sparsity for the level two lookup.
}
\vspace{2mm}
\label{tab:hierarchical-kernel}
\scriptsize
\begin{tabular}{c|c|c|c}
\toprule
L1/L2 Number of Centroids (as a Percentage of Context Length) &  Baseline Latency (1-Level) & Hierarchical Latency (L1 / L2) & Speedup  \\
\midrule
1\%/5\% Centroids & 1  & 0.82 (0.37 / 0.45) & 1.22 \\ 
0.5\%/5\% Centroids & 1 & 0.71 (0.27 / 0.44) & 1.41 \\
\bottomrule
\end{tabular}
\end{table*}

\begin{table*}[h!]
\centering
\caption{
Runtime (in minutes) for clustering with our one-level and hierarchical clustering method for the LWM-Text-Chat-1M model with 128K fixed context.
Note that clustering is performed \textit{offline} for the fixed context ahead of inference time, so it does not contribute to the inference-time latency for our method.
}
\vspace{2mm}
\label{tab:cluster-runtime}
\scriptsize
\begin{tabular}{c|c|c}
\toprule
Configuration &  One-Level Clustering (5\% Centroids) & Hierarchical Clustering (1\% / 5\% Centroids) \\
\midrule
Latency (minutes) &  23 & 24\\
\bottomrule
\end{tabular}
\end{table*}

\section{Clustering Runtime}
\label{appendix:cluster-runtime}

Table \ref{tab:cluster-runtime} shows the clustering runtime for our method with 128K context length for both our one-level and hierarchical (two-level) clustering approaches.
The clustering runtime was measured using the LWM-Text-Chat-1M model on 8 NVIDIA H100 NVL GPUs, and it includes the time for collecting the KV cache, performing clustering, and calibrating for the global threshold.
The clustering runtime for both the single-level and hierarchical case is 23-24 minutes.
Note that clustering is a one-time cost which is performed offline and therefore does not contribute to inference latency.

\end{document}